%% file: main.tex
\documentclass[11pt,a4paper]{article}

\usepackage[utf8]{inputenc}
\usepackage[T1]{fontenc}
\usepackage{amsmath,amssymb,amsfonts}
\usepackage{graphicx}
\usepackage{booktabs}
\usepackage{hyperref}
\usepackage{url}
\usepackage{xcolor}
\usepackage{natbib}
\usepackage{enumitem}
\usepackage{multirow}
\usepackage{caption}
\usepackage{subcaption}
\usepackage[margin=1in]{geometry}
\usepackage{listings}
\usepackage{tikz}
\usetikzlibrary{calc,positioning,arrows.meta}

\hypersetup{
    colorlinks=true,
    linkcolor=blue,
    citecolor=blue,
    urlcolor=blue
}

\title{Forage V2: Knowledge Evolution and Transfer \\ in Autonomous Agent Organizations}

\author{
    Huaqing Xie \\
    Independent Researcher \\
    \texttt{xhq422986742@gmail.com}
}

\date{}

\begin{document}
\maketitle

\begin{abstract}
Autonomous agents operating in open-world tasks---where the completion boundary is not given in advance---face a structural challenge we term \emph{denominator blindness}: they systematically underestimate the scope of the target space, reporting high coverage against a fraction of what exists. Forage V1 addressed this for individual runs through co-evolving evaluation (an independent Evaluator discovers and refines what ``complete'' means) and method isolation (the Evaluator and Planner cannot see each other's code). V2 extends the architecture from a single expedition to a \textbf{learning organization}: experience accumulates across runs, transfers across model capabilities, and institutional safeguards prevent the degradation of accumulated knowledge.

We demonstrate two claims across three task types (web scraping, API queries, mathematical reasoning). \textbf{Knowledge accumulation}: over six runs, knowledge entries grow from 0 to 54 (NVIDIA task), and the Evaluator's denominator estimates stabilize as domain understanding deepens. \textbf{Knowledge transfer}: a weaker agent (Sonnet) seeded with a stronger agent's (Opus) accumulated knowledge narrows a 6.6 percentage-point coverage gap to 1.1pp, halves cost (\$9.40$\to$\$5.13), converges in half the rounds (mean 4.5 vs.\ 7.0), and three independent seeded runs arrive at exactly the same denominator estimate (266), suggesting that organizational knowledge calibrates evaluation itself.

V2's contribution is architectural, not algorithmic: rather than making one model stronger through fine-tuning or reinforcement learning, it designs institutions---audit separation, contract protocols, organizational memory---that make any agent more reliable upon entry. The accumulated experience is organizational, model-agnostic, and transferable: stored as readable documents where each run enriches the institution, and every future agent---regardless of provider or capability level---inherits that enrichment. The deeper claim is that a minimal set of structural constraints enables credible autonomous judgment in spaces where no ground truth is given.
\end{abstract}

\input{sections/introduction}
\input{sections/background}
\input{sections/v2_method}
\input{sections/experiments}
\input{sections/analysis}
\input{sections/discussion}
\input{sections/related}
\input{sections/conclusion}

\input{sections/appendix}

\bibliographystyle{plainnat}
\bibliography{references}

\end{document}

%% file: sections/introduction.tex
\section{Introduction}
\label{sec:intro}

\subsection{The Harness Era}
\label{sec:intro:harness}

The development of autonomous AI agents has followed a three-stage progression. \textbf{Pre-training} builds foundational capabilities through large-scale learning on text corpora. \textbf{Post-training}---reinforcement learning from human feedback, test-time compute scaling, and related techniques---refines these capabilities toward human preferences and reasoning depth. The third stage---variously called the \textbf{harness layer}, scaffolding, or orchestration layer---equips agents with the external structure needed to operate in the real world: retrieval-augmented generation, persistent memory, tool use, skill composition, and workflow orchestration.

The harness layer is where much of today's applied AI engineering happens---Claude Code \citep{anthropic2025claude}, Cursor, and similar systems represent practical instantiations. Yet a critical dimension remains underserved: \textbf{evaluation}. Who judges whether the agent did enough? When should it stop?

Current approaches fall into two categories. The first relies on \textbf{human-authored evaluation criteria}: a person defines what ``complete'' means before the agent begins, and the agent is measured against this fixed standard \citep{anthropic2025swebench}. This works for benchmarks with known answers but breaks down in open-world tasks where the target space is itself unknown. In practice, the gap is filled by \textbf{humans in the loop}: users repeatedly correct agents that declare premature completion, redirect exploration that missed obvious sources, and manually judge when the work is ``good enough.'' This pattern---the human as implicit evaluator---is the default in current agent systems, and it does not scale. The second approach lets the \textbf{agent evaluate itself}---and this is where a systematic failure mode emerges.

\subsection{V1: Denominator Blindness and Co-Evolving Evaluation}
\label{sec:intro:v1}

In prior work \citep{foragev1}, we identified and named \textbf{denominator blindness}: the systematic tendency of autonomous agents to report high task completion against an underestimated total. An agent tasked with exploring an open-ended space---collecting all articles from a publication archive, or gathering all products in a category---may find a fraction of the target items, declare its job done, and report 100\% coverage. The agent's numerator may be accurate, but it never discovered the denominator.

Forage V1 addressed this for individual runs through two mechanisms. First, \textbf{co-evolving evaluation}: rather than fixing evaluation criteria before execution, V1 employs two LLM agents---an Evaluator that discovers and refines the definition of ``complete,'' and a Planner that designs and executes strategies. Both evolve simultaneously: as the Planner discovers new information, the Evaluator revises its assessment of the total scope; as the Evaluator identifies gaps, the Planner targets them. A non-LLM Executor runs the Planner's scripts deterministically. Second, \textbf{method isolation}: the Evaluator and Planner cannot see each other's code. The Evaluator writes \texttt{eval.py} (how to measure completeness); the Planner writes \texttt{action.py} (how to execute). Neither can read the other's script. This prevents self-serving evaluation---the structural tendency for an agent grading its own work to conflate ``I tried hard'' with ``I succeeded.''

V1 demonstrated that this architecture works across two benchmarks---government announcement archives and a historical publication collection. Without independent evaluation, a single agent self-reported 100\% coverage at 15.9\% actual recall; with full Forage, agents achieved 98.8\% actual recall with calibrated self-assessment. The architecture transferred across two task types without code changes.

\subsection{From Expedition to Organization}
\label{sec:intro:org}

V1 operates as a single expedition: a team ventures into unknown territory, draws a map, and returns. The map reflects the team's best understanding---credible but not necessarily complete, shaped by the sources discovered and the definitions adopted along the way. The next team, however, starts from scratch. Hard-won discoveries about the territory---which paths lead nowhere, which sources are reliable, what the boundaries look like---are lost between expeditions.

V2 extends Forage from a single expedition to a \textbf{learning organization}. After each run, both agents independently extract transferable lessons through a post-mortem process. These lessons accumulate in a persistent knowledge base, and subsequent runs begin with this accumulated context. The organization remembers.

This is not merely a technical improvement---it reflects a philosophical shift. V1 asked: \emph{how can two agents establish credible completion judgments?} V2 asks: \emph{how can an agent organization accumulate and transfer experience while maintaining the integrity guarantees that make its judgments trustworthy?}

The answer draws from institutional design rather than agent enhancement: method isolation as audit separation, format contracts as public agreements, and knowledge as organizational memory that newcomers consult but are not bound by (\S\ref{sec:discussion:institutional}). Forage does not make individual agents stronger; it designs institutions that make ordinary agents reliable.

\subsection{Contributions}
\label{sec:intro:contributions}

This report presents two claims, supported by experiments across three task types:

\begin{enumerate}[leftmargin=*]
    \item \textbf{Knowledge Accumulation.} The organization learns across runs. Knowledge entries grow from 0 to 54 over six runs (NVIDIA task), and the Evaluator's denominator estimates stabilize as accumulated domain knowledge narrows the search space.

    \item \textbf{Knowledge Transfer.} A stronger agent's (Opus) accumulated knowledge enables a weaker agent (Sonnet) to approach comparable performance. Without knowledge, Sonnet achieves 90.7--96.5\% coverage in 6--8 rounds (mean 93.1\%). With Opus's accumulated knowledge, Sonnet achieves 92.5--100\% coverage in 3--7 rounds (mean 98.6\%, with 5 of 6 runs at 99.7--100\%) at half the cost---and its denominator estimates converge to the same range as Opus's.
\end{enumerate}

Both claims are demonstrated on NVIDIA desktop GPU data collection (web scraping, 265--411 candidate products across conditions) and supported by UniProt protein collection (API queries, 28--30 candidates) and Q10 mathematical reasoning (First Proof benchmark). Cross-task knowledge transfer---whether lessons from one task domain benefit a different domain---is not claimed; this is V3 future work. (We note that some knowledge entries are obviously universal---e.g., ``this source blocks automated access''---but we do not measure cross-domain transfer in this report.)

The architecture's working unit is minimal: two LLM agents (Evaluator + Planner) and one deterministic Executor. This simplicity is deliberate---a minimal set of structural constraints that produces reliable autonomous judgment. The architecture composes naturally: additional agents, evaluation dimensions, or knowledge scopes can be layered without changing the core protocol.

The remainder of this report is organized as follows. Section~\ref{sec:background} reviews the V1 architecture and formalizes the problem. Section~\ref{sec:v2method} describes V2's extensions: knowledge evolution, knowledge transfer, and architectural improvements. Section~\ref{sec:experiments} presents the experimental design and results. Section~\ref{sec:analysis} analyzes behavioral patterns and anomalies. Section~\ref{sec:discussion} discusses limitations, the institutional design perspective, and future directions. Section~\ref{sec:related} positions Forage relative to existing work. Section~\ref{sec:conclusion} concludes.

%% file: sections/background.tex
\section{Background \& V1 Foundation}
\label{sec:background}

This section formalizes the problem V1 solved and summarizes the V1 architecture, establishing the foundation on which V2 builds. Readers familiar with \citet{foragev1} may skim to Section~\ref{sec:background:open}.

\subsection{Problem Formulation: Open-World Tasks}
\label{sec:background:problem}

We define an \textbf{open-world task} as one where the completion boundary is not given in advance. The agent receives a task specification $\mathcal{T}$ (e.g., ``collect all products in this category,'' ``find all proteins associated with this disease,'' or ``prove this theorem and verify all edge cases'') and must produce an output $\mathcal{O}$ that satisfies an unknown completeness criterion $\mathcal{C}$. The criterion $\mathcal{C}$ includes both the \emph{denominator}---the total scope of target items---and the \emph{quality standard}---what constitutes an acceptable result.

In contrast to closed-world tasks (benchmarks, unit tests, competitions) where $\mathcal{C}$ is provided, open-world tasks require the agent to \emph{discover} $\mathcal{C}$ as part of the work itself. The agent faces a dual search: exploring the space to produce results (expanding the numerator) while simultaneously searching for the boundary of what exists (discovering the denominator).

\paragraph{Denominator blindness.} When an agent evaluates its own completeness, it estimates both numerator and denominator. We observed that agents systematically underestimate the denominator: in V1 experiments on a government announcement archive and a historical publication collection, a single agent self-reported 100\% coverage at 15.9\% actual recall \citep{foragev1}. The agent's error was not in execution but in scope estimation---it never searched broadly enough to discover the full extent of the target space. We term this \textbf{denominator blindness}: the structural tendency of self-evaluating agents to conflate ``everything I found'' with ``everything that exists.''

Denominator blindness is not a prompting failure or a model limitation. It is a structural consequence of combining the roles of executor and evaluator in a single agent. Anthropic has observed the same phenomenon, noting that agents ``confidently praise mediocre work'' when evaluating their own output \citep{anthropic2025swebench}. ByteDance's WideSearch measured the problem at scale, finding that autonomous agents achieve only 5.1\% actual coverage while self-reporting high confidence \citep{widesearch2025}.

\subsection{The Forage V1 Architecture}
\label{sec:background:v1arch}

V1 addresses denominator blindness through structural separation rather than better prompting. The architecture comprises three components operating in an iterative loop:

\paragraph{Evaluator Agent.} The Evaluator is responsible for answering: \emph{what does ``complete'' mean?} In Round~1, it explores the data source landscape---examining sitemaps, API endpoints, archival indices---and establishes an initial denominator estimate anchored to verifiable external sources. It writes \texttt{eval.py}, a deterministic Python script that computes coverage metrics. In subsequent rounds, it audits the latest results, may revise the denominator based on new evidence, and decides whether to continue or stop.

\paragraph{Planner Agent.} The Planner is responsible for maximizing the numerator. It receives coverage metrics, a gap report, and a discovery summary from the Evaluator (what sources exist, not how they were found). It designs a collection strategy and writes \texttt{action.py}, an executable script. The Planner has full autonomy over implementation---which libraries, concurrency model, rate limiting, and source prioritization.

\paragraph{Executor.} The Executor is a non-LLM deterministic process that runs the Planner's script and then runs the Evaluator's script to produce updated metrics. It is not an agent---it performs no reasoning, makes no decisions, and generates no code.

Each round proceeds as: Evaluator $\to$ Planner $\to$ Executor $\to$ (metrics feed next round). The loop terminates when the Evaluator decides coverage is sufficient, when a budget is exhausted, or when a plateau is detected.

\paragraph{Method isolation.} The Evaluator and Planner operate as independent entities whose internal methods are mutually invisible, enforced through four layers:
\begin{enumerate}[leftmargin=*]
    \item \textbf{Method isolation}: each agent's code is hidden from the other.
    \item \textbf{Context isolation}: separate LLM invocations with independent contexts.
    \item \textbf{Temporal separation}: the Evaluator defines the standard before the Planner executes against it.
    \item \textbf{Independent anchoring}: the denominator must be grounded independently of execution results---in verifiable external sources for data tasks (sitemaps, API counts, archival indices), or in the logical structure of the problem itself for reasoning tasks (what dimensions of correctness must be verified).
\end{enumerate}

Without method isolation, cognitive anchoring suppresses independent exploration. If the Evaluator sees that the Planner uses table-of-contents pages, it may limit its denominator search to what those pages reveal. If the Planner sees that the Evaluator defines a denominator of 110, it may stop at 110 items. Isolation frees both agents to independently expand their cognitive boundaries.

\paragraph{Stopping criteria.} Forage employs a dual-layer stopping mechanism: an \emph{objective layer} (deterministic metrics from \texttt{eval.py}) and a \emph{judgment layer} (the Evaluator Agent interprets metrics in context---has the denominator stabilized? are there remaining avenues? is continued work likely to yield diminishing returns?). A hard budget (\texttt{max\_rounds}) serves as a safety fallback.

\subsection{What V1 Left Open}
\label{sec:background:open}

V1 demonstrated that co-evolving evaluation mitigates denominator blindness within a single run. Three limitations motivated V2:

\begin{enumerate}[leftmargin=*]
    \item \textbf{No experience accumulation.} Each run starts from zero. The Evaluator must rediscover the scope of the target space; the Planner must reinvent strategies from scratch. Hard-won lessons about which sources are accessible, what pitfalls exist, and how the domain is structured are lost between expeditions.

    \item \textbf{No knowledge transfer.} One expedition's discoveries cannot benefit the next. If a team spends six runs mapping a domain---discovering which sources are reliable, what the boundary conditions are, how to avoid known pitfalls---the next team must rediscover all of this from scratch. This is unlike any successful human knowledge enterprise: scientific journals, field manuals, and institutional handbooks exist precisely so that later expeditions inherit earlier ones' hard-won understanding, regardless of team composition or capability.

    \item \textbf{Fragile isolation.} V1 enforced method isolation by hiding files with a dotfile prefix (\texttt{.eval.py}). Post-hoc analysis of V1 experiment logs confirmed no breach occurred during the original V1 study. However, during early V2 experiments using the same mechanism, a CLI log audit revealed that an agent discovered and directly accessed the hidden file---demonstrating that convention-level isolation is inherently fragile and will eventually be defeated by sufficiently capable agents (see \S\ref{sec:analysis:breach}).
\end{enumerate}

V2 addresses all three: knowledge evolution (\S\ref{sec:v2:knowledge}) for accumulation, a transfer protocol (\S\ref{sec:v2:transfer}) for cross-capability inheritance, and physical workspace isolation (\S\ref{sec:v2:arch:isolation}) for robust separation.

%% file: sections/v2_method.tex
\section{V2: From Expedition to Organization}
\label{sec:v2method}

Forage V1 demonstrated that co-evolving evaluation enables autonomous agents to establish credible completion judgments in open-world tasks. However, each V1 run starts from zero---the Evaluator must rediscover the data landscape, the Planner must reinvent collection strategies, and hard-won insights about source reliability or parsing pitfalls are lost between runs. V2 extends the architecture from a single expedition to a \textbf{learning organization}: experience accumulates across runs, transfers to new agents, and institutional safeguards prevent the degradation of accumulated knowledge.

This section describes the three extensions that constitute V2: a knowledge evolution mechanism that captures and delivers cross-run experience (\S\ref{sec:v2:knowledge}), a knowledge transfer protocol that enables weaker agents to inherit stronger agents' accumulated experience (\S\ref{sec:v2:transfer}), and architectural improvements that strengthen the isolation guarantees required for trustworthy knowledge (\S\ref{sec:v2:arch}). Figure~\ref{fig:architecture} provides a conceptual overview. We include engineering details where they illuminate design decisions; readers interested primarily in empirical results may proceed to Section~\ref{sec:experiments}.

\input{figures/architecture_v2}

\subsection{Knowledge Evolution}
\label{sec:v2:knowledge}

The central extension in V2 is a mechanism for agents to extract, store, and inherit experience across runs. We call this \emph{knowledge evolution}: the organization's collective understanding grows with each completed run, and future runs begin with this accumulated context.

\paragraph{Post-mortem extraction.} After each run completes (either by the Evaluator's stop decision or by budget exhaustion), both agents independently extract transferable lessons from the run's trajectory. The Evaluator receives a narrative of its own rounds---denominator estimates, coverage trajectories, discovery summaries---but not the Planner's strategies or scripts. Symmetrically, the Planner receives its own strategy history and collection results, but not the Evaluator's evaluation logic. This preserves method isolation even during knowledge extraction.

Each agent produces a set of structured lessons:
\begin{lstlisting}[basicstyle=\small\ttfamily, frame=single, xleftmargin=2em]
{"id": "wikipedia_table_parsing_fragile",
 "scope": "web_scraping",
 "type": "advisory",
 "summary": "Wikipedia table headers vary by series",
 "content": "The GeForce RTX 30 series uses ..."}
\end{lstlisting}

Lessons are categorized by scope (universal, domain-specific, or task-specific) and stored as individual markdown files with YAML frontmatter under a persistent \texttt{knowledge/} directory.

\paragraph{Append-only versioning.} A critical design decision is that knowledge entries are \emph{never overwritten}. When an entry with identifier \texttt{X} already exists, a new observation is stored as \texttt{X\_v2}, \texttt{X\_v3}, and so on. This preserves the full history of the organization's evolving understanding. A single run's observation is a new data point, not a refutation of prior wisdom---reconciliation is left to future curation (see \S\ref{sec:discussion:vision}).

\paragraph{Delivery via system prompt.} At the start of each run, the harness stages accumulated knowledge files into the shared workspace and generates an \texttt{INDEX.md}---a structured catalog organized by scope. This index is injected into each agent's system prompt, giving both the Evaluator and the Planner access to the organization's accumulated experience from the first turn onward.

\paragraph{Advisory, not prescriptive.} Knowledge entries carry a \texttt{type: advisory} designation. They represent the organization's experience, not instructions. The receiving agent decides what to adopt based on its own assessment of the current situation. This design acknowledges that knowledge can become stale or context-dependent: a parsing strategy that worked for Wikipedia in one run may not apply when TechPowerUp becomes accessible. We observe in our experiments that agents selectively adopt knowledge---sometimes following prior advice closely, sometimes diverging based on their own exploration (\S\ref{sec:analysis}).

\subsection{Knowledge Transfer}
\label{sec:v2:transfer}

Knowledge evolution accumulates experience within a single agent lineage (e.g., six consecutive Opus runs). Knowledge \emph{transfer} tests whether this accumulated experience benefits a different agent---specifically, whether a weaker model can leverage a stronger model's organizational experience.

\paragraph{Transfer protocol.} The transfer mechanism is straightforward: the knowledge directory accumulated by the strong agent (Opus) across its runs is provided as the starting knowledge for the weak agent's (Sonnet) runs. No modification is made to the knowledge files---the weak agent receives exactly the same \texttt{INDEX.md} and lesson files that the strong agent would see in its next run. The transfer operates at the level of organizational documents, not model weights or fine-tuning data.

\paragraph{The eval\_contract.md protocol.} A key interface in knowledge transfer is \texttt{eval\_contract.md}, a document the Evaluator writes to the shared workspace each round. This contract specifies the expected output format, field definitions, and evaluation dimensions---effectively, the ``terms of engagement'' that the Planner must satisfy. In the seeded condition, the Evaluator inherits knowledge about what constitutes a complete product definition (e.g., which variant types to count, which fields are mandatory), enabling it to write a more precise contract from Round~1. The Planner, reading this contract alongside accumulated strategic knowledge, can begin with a curated baseline rather than exploratory scraping.

\paragraph{What transfers and what doesn't.} The transferred knowledge includes: domain understanding (e.g., ``NVIDIA desktop GPUs span 24 product series from 1995--2024''), source reliability (e.g., ``TechPowerUp blocks automated access''), quality lessons (e.g., ``Wikipedia table column alignment varies by series generation''), and strategic patterns (e.g., ``curated hardcoded baselines avoid parsing errors''). It does not include: the Evaluator's \texttt{eval.py} code, the Planner's \texttt{action.py} code, or any session-specific context. Method isolation is maintained across the transfer boundary.

\paragraph{Why strong$\to$weak transfer works.} A stronger model produces richer, more articulate knowledge entries---not because it writes better prose, but because it encounters more of the problem space within its turn budget and can distill observations at a level of abstraction accessible to any reader. The weaker model, receiving these distilled lessons, skips the exploration that produced them. This asymmetry is by design: accumulated knowledge is organizational and model-agnostic, so the receiving agent need not match the generating agent's capability to benefit from its conclusions. Whether knowledge quality degrades when generated by weaker models (and whether a curation layer can mitigate this) is V3 future work.

\subsection{Architectural Improvements}
\label{sec:v2:arch}

V2 introduces several architectural improvements that strengthen the foundation on which knowledge evolution operates. We describe the design principles; implementation details are available in the codebase.

\subsubsection{Hardened Method Isolation}
\label{sec:v2:arch:isolation}

Any system that grants an autonomous agent filesystem access while relying on naming conventions to hide sensitive files is inherently fragile---a sufficiently capable agent will eventually discover hidden paths through exploration, error messages, or directory enumeration. V1 enforced method isolation by hiding files before invoking the opposing agent. During V2 development, a CLI log audit revealed that this mechanism was defeated: an agent discovered and directly accessed the opposing agent's hidden script (see \S\ref{sec:analysis:breach}). The resulting ``peeked'' knowledge was then extracted as legitimate experience during post-mortem, contaminating the knowledge base for subsequent runs.

V2 replaces this with \textbf{physical workspace separation}: each agent operates in its own private directory, with access to shared resources (results, metrics, knowledge, the format contract) through a common interface. The opposing agent's code does not exist in the filesystem namespace visible to the current agent---a stronger guarantee than filename-based hiding. The key principle is that isolation must be enforced at the infrastructure level, not the convention level, especially when accumulated knowledge creates incentives for shortcutting.

\subsubsection{Session Persistence and Recovery}
\label{sec:v2:arch:session}

V1 agents had no memory across rounds within a run. V2 maintains agent context across rounds through persistent sessions: the agent retains its reasoning history, denominator revisions, and strategic memory as a continuous narrative rather than reconstructing context from logs each round.

Agent sessions can fail due to API errors, timeouts, or context exhaustion. V2 implements a recovery mechanism with two stages: first, the harness attempts to salvage usable work the agent may have written to disk despite the session failure; if salvage fails, a replacement agent is deployed with a role-appropriate trajectory summary as context, preserving method isolation in the handoff. Each agent maintains an independent session---recovering one does not affect the other.

The 8-round budget reflects an engineering constraint: persistent sessions accumulate context across rounds, and current LLM context windows impose a practical ceiling on how many rounds a single session can sustain before degradation. Eight rounds provides sufficient iteration depth for knowledge accumulation to matter while remaining within context limits. The interaction between round budgets, context windows, and session management is discussed further in \S\ref{sec:discussion:limits}.

\subsubsection{Exploration Drive}
\label{sec:v2:arch:selfaudit}

A learning organization faces a paradox: accumulated knowledge improves efficiency but can also induce premature convergence. If the Evaluator ``knows'' the denominator is approximately 270, it may accept that number without independent verification. V2 counters this through \emph{exploration drive}---mechanisms at three architectural levels that resist premature satisfaction:

\begin{enumerate}[leftmargin=*]
    \item \textbf{Prompt level (self-audit checklist).} Every Evaluator round (Round~2 onward) includes a fixed set of self-challenge questions: Is my denominator still accurate? Is my evaluation rigorous enough? Am I being too lenient? Should I expand the denominator rather than accepting overcounting? These do not prescribe answers---they force the Evaluator to confront potential blind spots before each stop/continue decision.
    \item \textbf{Context level (persistent sessions).} Because the Evaluator retains its full reasoning history across rounds, it can observe its own trajectory: ``I estimated 255 in Round~1, then 273 in Round~3---am I still discovering, or have I converged?'' This self-awareness enables calibrated confidence rather than blind acceptance of the latest estimate.
    \item \textbf{Harness level (round budget).} The multi-round structure itself is an exploration mechanism: even if the Evaluator is satisfied, the budget allows the Planner additional rounds to surface new data that might challenge the current denominator. The 8-round cap ensures this exploration has a cost ceiling.
\end{enumerate}

\subsubsection{Runtime Parameters as Supplies}
\label{sec:v2:arch:params}

Extending the expedition metaphor: before each run, the harness provisions the team with \emph{supplies}---runtime constraints that determine how much work the agents can do per sortie. V2 fixes three interlocking supply parameters: a per-round \emph{turn budget} (\texttt{max\_turns}) that limits how many tool-use actions an agent can take in a single round, a per-invocation \emph{timeout} that prevents indefinite hangs, and a \emph{maximum round count} that bounds total expedition length. Each ``turn'' is one agentic cycle: the model reasons, invokes tools (filesystem operations, web requests, code execution), and receives results. A budget of 15 turns means 15 such cycles before the agent must stop and return control to the harness.

The turn budget serves a dual role: it is both a \emph{supply} (the agent's action capacity per round) and a \emph{constraint} (forcing frequent inter-round pauses, where the Evaluator can reassess and knowledge can accumulate). This duality is deliberate---lower turn budgets make each round shallower, but produce more inter-round checkpoints, making knowledge evolution more consequential. These constraints interlock: turns control depth per round, timeouts prevent indefinite hangs, and round limits bound total cost.

Different tasks require different supply profiles. Structured exploration tasks (NVIDIA, UniProt) use modest per-round budgets (\texttt{max\_turns=15}) that force frequent inter-round pauses---each pause is an opportunity for the Evaluator to reassess and for knowledge to accumulate. Reasoning tasks (Q10) require deeper per-round engagement (\texttt{max\_turns=50}) because mathematical derivations cannot be meaningfully interrupted mid-proof. The Q6 results (\S\ref{sec:exp:q6}) suggest that when supplies are insufficient for the task's cognitive demands, knowledge accumulation alone cannot compensate---a finding that motivates V3's dynamic supply allocation.

%% file: figures/architecture_v2.tex
\begin{figure}[t]
\centering
\begin{tikzpicture}[
    node distance=1.2cm and 1.8cm,
    box/.style={draw, rounded corners, minimum height=0.8cm, minimum width=1.6cm, font=\small},
    agent/.style={box, fill=blue!8},
    shared/.style={box, fill=gray!12},
    knowledge/.style={box, fill=green!10, minimum width=2.2cm},
    arrow/.style={->, >=stealth, thick},
    dasharrow/.style={->, >=stealth, dashed, thick},
    label/.style={font=\scriptsize, text=gray!70!black},
]

\node[agent] (eval) {Evaluator};
\node[shared, right=2.5cm of eval] (shared) {Shared};
\node[agent, right=2.5cm of shared] (plan) {Planner};

\draw[thick, red!60, dashed] ($(eval.north west)+(-0.3,0.3)$) rectangle ($(eval.south east)+(0.3,-0.3)$);
\draw[thick, red!60, dashed] ($(plan.north west)+(-0.3,0.3)$) rectangle ($(plan.south east)+(0.3,-0.3)$);

\node[label, above=0.05cm of eval] {eval\_ws (private)};
\node[label, above=0.05cm of shared] {dataset, metrics, contract};
\node[label, above=0.05cm of plan] {plan\_ws (private)};

\draw[arrow] (eval) -- node[above, font=\scriptsize] {eval\_contract} (shared);
\draw[arrow] (shared) -- node[above, font=\scriptsize] {metrics} (plan);
\draw[arrow, bend left=20] (plan) to node[below, font=\scriptsize] {dataset} (shared);
\draw[arrow, bend left=20] (shared) to node[below, font=\scriptsize] {coverage} (eval);

\node[font=\large, red!70] at ($(eval)!0.5!(plan)+(0,-0.9)$) {\texttimes\ no mutual visibility};

\node[knowledge, below=2.2cm of shared] (kb) {Knowledge Base};

\node[box, fill=blue!5, below left=1.5cm and 1.5cm of kb, minimum width=1.2cm] (r1) {Run 1};
\node[box, fill=blue!5, right=0.6cm of r1, minimum width=1.2cm] (r2) {Run 2};
\node[right=0.3cm of r2, font=\small] (dots) {\ldots};
\node[box, fill=blue!5, right=0.3cm of dots, minimum width=1.2cm] (rn) {Run $N$};

\draw[arrow, blue!70] (r1.north) -- ++(0,0.4) -| node[pos=0.25, above, font=\scriptsize] {post-mortem} (kb.south west);
\draw[arrow, blue!70] (r2.north) -- ++(0,0.6) -| (kb.south);

\draw[dasharrow, green!60!black] (kb.south east) |- node[pos=0.75, right, font=\scriptsize] {seed} (rn.north);

\node[box, fill=green!8, below right=1.5cm and 1.5cm of kb, minimum width=2cm] (transfer) {Sonnet (seeded)};
\draw[dasharrow, green!60!black] (kb.east) -| node[pos=0.3, above, font=\scriptsize] {transfer} (transfer.north);

\node[font=\footnotesize\bfseries, anchor=west] at ($(eval.north west)+(-0.8, 0.7)$) {Within a Run};
\node[font=\footnotesize\bfseries, anchor=west] at ($(r1.north west)+(-0.5, 1.0)$) {Across Runs};

\end{tikzpicture}
\caption{Forage V2 architecture. \textbf{Top:} within a single run, the Evaluator and Planner operate in isolated workspaces with no mutual visibility; they coordinate through shared artifacts (dataset, metrics, eval\_contract). \textbf{Bottom:} across runs, each run's post-mortem extracts knowledge into a persistent base. Subsequent runs---and weaker models via transfer---inherit accumulated organizational experience.}
\label{fig:architecture}
\end{figure}

%% file: sections/experiments.tex
\section{Experiments}
\label{sec:experiments}

\subsection{Task Descriptions}
\label{sec:exp:tasks}

We evaluate Forage V2 on three tasks spanning different exploration modalities and cognitive demands:

\paragraph{NVIDIA Desktop GPU Collection (web scraping).} Collect specifications for all NVIDIA desktop GPU products from 1990 to 2024. Each record requires nine fields: product name, chip name, release date, bus interface, memory capacity, memory type, memory bit width, core frequency, and memory frequency. The data source landscape includes Wikipedia (accessible, structured tables with varying formats across GPU generations), TechPowerUp (comprehensive but bot-protected), and NVIDIA's own product archive. The task is challenging because ``product'' is ambiguous: does a memory variant (DDR3 vs GDDR5) count as a separate product? Does an OEM rebadge? The true denominator is unknown and must be discovered.

\paragraph{UniProt T2D Protein Collection (API queries).} Collect all reviewed (Swiss-Prot) protein entries associated with Type 2 Diabetes from the UniProt database. This task uses the UniProt REST API with structured queries. The denominator is relatively well-defined (reviewed entries matching disease annotations), making it a useful complement to the ambiguous NVIDIA task. Records require gene name, protein name, organism, function, and disease association fields.

\paragraph{Q10 Mathematical Reasoning (First Proof benchmark).} Answer Question 10 from the First Proof benchmark \citep{firstproof2026}---a set of ten previously unpublished, research-level mathematics questions with encrypted answers. Q10, posed by Tammy Kolda, concerns efficient iterative solvers for RKHS-constrained tensor decomposition (full statement in Appendix~\ref{app:tasks:math}). Published answers are not available; our purpose is not to verify mathematical correctness against a known solution, but to observe whether the Evaluator autonomously discovers and refines the space of verification dimensions. Unlike the structured exploration tasks, there is no ``denominator'' in the traditional sense. The denominator here is the number of evaluation criteria the Evaluator defines (what aspects of the solution must be checked?), and co-evolution manifests as the Evaluator expanding its test suite while the Planner refines the solution. We additionally report Question~6 from the same benchmark---a substantially harder problem---as a case study in difficulty frontiers (\S\ref{sec:exp:q6}).

\subsection{Experimental Conditions}
\label{sec:exp:conditions}

For NVIDIA and Q10, we run three conditions; for UniProt, one condition:

\begin{table}[ht]
\centering
\small
\begin{tabular}{llp{7.5cm}}
\toprule
\textbf{Condition} & \textbf{Model} & \textbf{Description} \\
\midrule
Opus & claude-opus-4-6 & 6 runs with knowledge accumulating from zero. Establishes the knowledge base. \\
Sonnet cold & claude-sonnet-4-6 & 6 runs with no prior knowledge. Weak agent baseline. \\
Sonnet seeded & claude-sonnet-4-6 & 6 runs carrying Opus's accumulated knowledge. Tests knowledge transfer. \\
\bottomrule
\end{tabular}
\caption{Experimental conditions. Within each task type, all conditions use identical task specifications and runtime parameters. Model versions are pinned to prevent silent upgrades.}
\label{tab:conditions}
\end{table}

Runtime parameters differ by task type. Structured exploration tasks (NVIDIA, UniProt) use \texttt{max\_turns=15}, \texttt{effort=medium}, \texttt{timeout=1200s}, and \texttt{max\_rounds=8}. Reasoning tasks (Q10) use \texttt{max\_turns=50}, \texttt{effort=high}, and \texttt{max\_rounds=8}. The lower turn budget for structured exploration forces multi-round iteration, making knowledge accumulation more consequential.

Each condition consists of 6 independent runs, producing 6 trajectory files with per-round data. All runs use physical workspace isolation.

\subsection{Claim 1: Knowledge Accumulation}
\label{sec:exp:claim1}

\begin{figure}[t]
\centering
\includegraphics[width=0.65\textwidth]{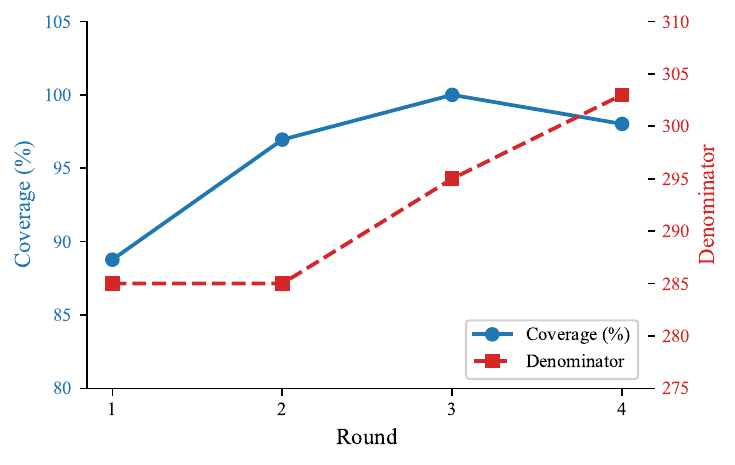}
\caption{Anatomy of a single run (NVIDIA Opus, run~002). Coverage and denominator co-evolve across four rounds: the Evaluator expands the denominator from 285 to 303 as it refines its product definition, causing a slight coverage dip in Round~4 despite no loss of collected records. This dynamic---the denominator growing alongside collection---is the co-evolution mechanism in action.}
\label{fig:single_run}
\end{figure}

\subsubsection{NVIDIA Opus: The Primary Evidence}

Table~\ref{tab:nvidia_opus} presents the six Opus runs. Several patterns are consistent across all runs:

\begin{table}[ht]
\centering
\small
\begin{tabular}{cccccc}
\toprule
\textbf{Run} & \textbf{Rounds} & \textbf{Final Denom} & \textbf{Coverage} & \textbf{Cost} & \textbf{Denom Trajectory} \\
\midrule
001 & 4 & 285 & 100.0\% & \$5.61 & 254$\to$280$\to$285$\to$285 \\
002 & 4 & 303 & 98.0\%  & \$5.72 & 285$\to$285$\to$295$\to$303 \\
003 & 4 & 274 & 100.0\% & \$4.59 & 295$\to$262$\to$274$\to$274 \\
004 & 4 & 269 & 100.0\% & \$5.63 & 256$\to$265$\to$268$\to$269 \\
005 & 4 & 273 & 100.0\% & \$4.36 & 255$\to$255$\to$273$\to$273 \\
006 & 4 & 265 & 100.0\% & \$4.38 & 252$\to$260$\to$265$\to$265 \\
\bottomrule
\end{tabular}
\caption{NVIDIA Opus: per-run results. All runs converge in exactly 4 rounds. Coverage is relative to each run's own final denominator.}
\label{tab:nvidia_opus}
\end{table}

\paragraph{Convergence stability.} All six runs converge in exactly 4 rounds---the minimum possible (explore, fill, fill, stop). No run hits the 8-round budget limit. Cost is stable: \$4.36--\$5.72 (mean \$5.05, std \$0.60).

\paragraph{Denominator co-evolution.} The final denominator varies from 265 to 303 across runs. This spread (14\%) reflects genuine ambiguity in what constitutes a ``product''---whether OEM rebadges, memory variants, and bus variants count as separate entries. Critically, each run's Evaluator arrives at its own definition independently and achieves near-complete coverage against that definition. Run~003 is notable: the Evaluator revised its denominator \emph{downward} (295$\to$262) after detecting overcounting, then corrected upward (262$\to$274)---demonstrating the self-audit mechanism in action.

\paragraph{Knowledge growth.} Knowledge entries accumulate monotonically: 0$\to$8$\to$22$\to$29$\to$38$\to$46$\to$54 across the six runs. Each run's post-mortem adds 6--14 new lessons covering domain knowledge (series structure, variant types), source reliability (TechPowerUp blocked, Wikipedia format variations), and quality patterns (deduplication rules, field validation).

\subsubsection{UniProt Opus: API Task Validation}

\begin{table}[ht]
\centering
\small
\begin{tabular}{ccccc}
\toprule
\textbf{Run} & \textbf{Rounds} & \textbf{Final Denom} & \textbf{Coverage} & \textbf{Cost} \\
\midrule
001 & 3 & 28 & 100.0\% & \$2.39 \\
002 & 4 & 28 & 100.0\% & \$2.81 \\
003 & 4 & 28 & 100.0\% & \$3.87 \\
004 & 4 & 28 & 100.0\% & \$3.76 \\
005 & 5 & 28 & 100.0\% & \$3.50 \\
006 & 3 & 30 & 99.9\%  & \$2.14 \\
\bottomrule
\end{tabular}
\caption{UniProt T2D Opus: per-run results. Denominator is highly stable (28 for 5/6 runs). Knowledge grows 0$\to$12$\to$18$\to$26$\to$35$\to$42.}
\label{tab:uniprot_opus}
\end{table}

\begin{figure}[t]
\centering
\includegraphics[width=0.7\textwidth]{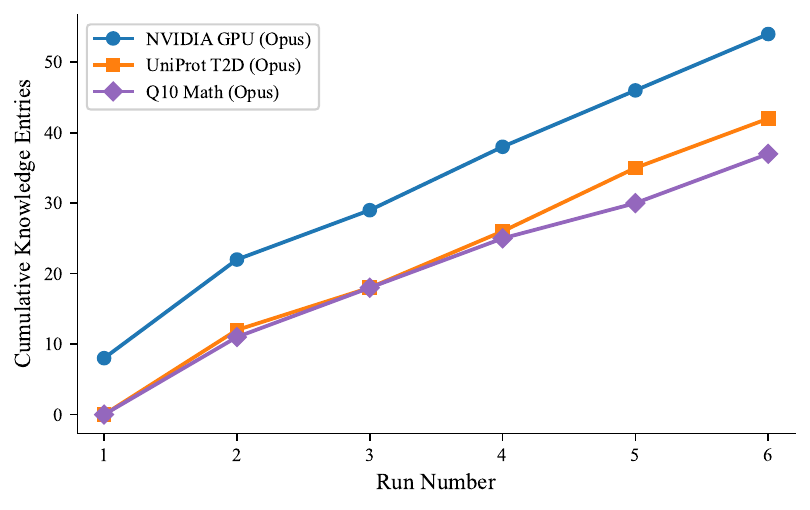}
\caption{Cumulative knowledge entries across six runs for three Opus conditions. All tasks show monotonic growth with consistent per-run increments (6--14 entries per run), indicating that post-mortem extraction reliably produces new lessons regardless of task domain.}
\label{fig:knowledge_growth}
\end{figure}

UniProt confirms that knowledge accumulation is not a single-task artifact. The denominator is far more stable than NVIDIA (28 vs 265--303), reflecting the well-defined nature of API queries. Knowledge entries accumulate 0$\to$42 across six runs, covering API query patterns, cross-validation strategies, and field extraction techniques.

\subsubsection{Q10 Opus: Reasoning Task Validation}

\begin{table}[ht]
\centering
\small
\begin{tabular}{ccccc}
\toprule
\textbf{Run} & \textbf{Rounds} & \textbf{Final Denom} & \textbf{Coverage} & \textbf{Cost} \\
\midrule
001 & 2 & 7 & 83.2\% & \$3.01 \\
002 & 2 & 7 & 100.0\% & \$2.48 \\
003 & 3 & 7 & 100.0\% & \$6.14 \\
004 & 2 & 7 & 100.0\% & \$2.85 \\
005 & 2 & 8 & 98.7\%  & \$1.97 \\
006 & 3 & 7 & 100.0\% & \$2.26 \\
\bottomrule
\end{tabular}
\caption{Q10 Opus: per-run results. The ``denominator'' here is the number of evaluation dimensions the Evaluator defines (7--8). Knowledge grows 0$\to$11$\to$18$\to$25$\to$30$\to$37.}
\label{tab:q10_opus}
\end{table}

Q10 validates that knowledge accumulation extends to reasoning tasks. The ``denominator'' in this context is the number of evaluation dimensions the Evaluator defines (e.g., correctness of bounds, verification of edge cases, completeness of proof steps)---typically 7--8 dimensions, self-discovered rather than externally enumerable. Run~001 achieves only 83.2\%; notably, the Evaluator stopped because it recognized the gap as an eval.py text-matching ceiling rather than a genuine content deficiency---a better stop decision than chasing evaluation artifacts to 100\%. Runs 002--006 achieve 98.7--100\%. Knowledge entries grow 0$\to$37, covering mathematical techniques, proof strategies, and evaluation rigor.

\subsection{Claim 2: Knowledge Transfer}
\label{sec:exp:claim2}

\subsubsection{NVIDIA: The Transfer Experiment}

\begin{figure}[t]
\centering
\includegraphics[width=\textwidth]{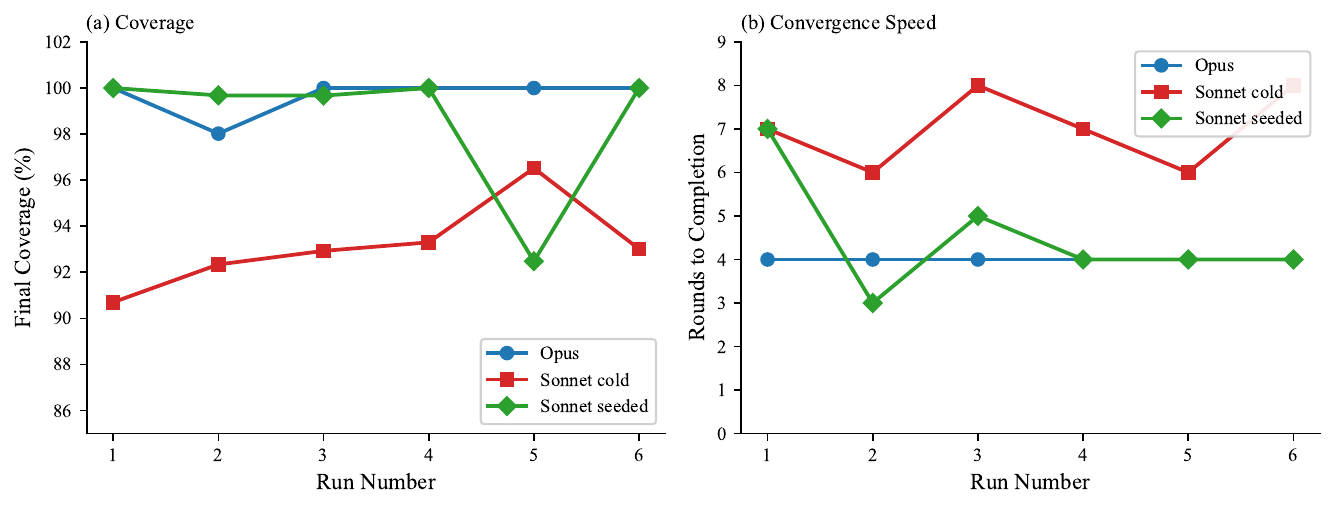}
\caption{NVIDIA knowledge transfer effect across six runs per condition. \textbf{(a)}~Seeded Sonnet's coverage approaches Opus and exceeds cold Sonnet on 5 of 6 runs. \textbf{(b)}~Seeded Sonnet converges in 3--7 rounds (mean 4.5, approaching Opus's 4.0), while cold Sonnet requires 6--8 rounds (mean 7.0).}
\label{fig:transfer}
\end{figure}

The NVIDIA task is the primary evidence for knowledge transfer because Sonnet cold \emph{visibly struggles} on it---unlike Q10, where Sonnet cold achieves 100\% on all runs (the task does not challenge Sonnet enough to reveal transfer effects). Table~\ref{tab:nvidia_comparison} presents the cross-condition comparison.

\begin{table}[ht]
\centering
\small
\begin{tabular}{lccc}
\toprule
\textbf{Metric} & \textbf{Opus} & \textbf{Sonnet Cold} & \textbf{Sonnet Seeded} \\
\midrule
Coverage (mean) & 99.7\% & 93.1\% & 98.6\% \\
Coverage (range) & 98.0--100\% & 90.7--96.5\% & 92.5--100\% \\
Cost (mean) & \$5.05 & \$9.40 & \$5.13 \\
Rounds (mean) & 4.0 & 7.0 & 4.5 \\
Denom range & 265--303 & 320--411 & 266--359 \\
\bottomrule
\end{tabular}
\caption{NVIDIA cross-condition comparison. Coverage is relative to each run's own denominator. Seeded Sonnet approaches Opus on all metrics while starting from the same model as cold Sonnet.}
\label{tab:nvidia_comparison}
\end{table}

\paragraph{Coverage improvement.} Cold Sonnet achieves 90.7--96.5\% coverage (mean 93.1\%). Seeded Sonnet achieves 92.5--100\% (mean 98.6\%), with 5 of 6 runs at 99.7--100\%. The single outlier (run\_005, 92.5\%) stopped when the Evaluator judged worst-case coverage (89.1\%) to be within the soft target's tolerance.

\paragraph{Efficiency improvement.} Cold Sonnet requires 6--8 rounds (mean 7.0); seeded Sonnet requires 3--7 rounds (mean 4.5), approaching Opus's consistent 4 rounds. Cost drops from \$9.40 to \$5.13---a 45\% reduction, matching Opus's \$5.05.

\paragraph{Denominator convergence---the strongest evidence.} Cold Sonnet's denominators range from 320 to 411 (spread of 91, or 28\%)---it consistently overestimates the product count due to imprecise deduplication. Seeded Sonnet's denominators start broad (run\_001: 359, the learning-curve run) but converge dramatically: runs 004, 005, and 006 all arrive at exactly 266---within Opus's range of 265--303. The Evaluator in run\_006 explicitly states: ``denominator 266 matches prior-run convergence confirmed in multiple independent knowledge-base entries.'' Knowledge transfer calibrates not only collection quality but evaluation quality.

\subsubsection{The Strategy Shift}

Beyond quantitative metrics, knowledge transfer produces a qualitative shift in approach. Cold Sonnet's strategies are reactive: it scrapes Wikipedia tables, encounters parsing bugs (column misalignment, process nodes parsed as chip names, memory frequency confused with shader counts), and spends multiple rounds fixing them. Strategy names reflect this reactive pattern---\emph{full parse}, \emph{fix and validate}, \emph{quality hardening}---each addressing bugs discovered in the previous round.

Seeded Sonnet's strategies are proactive. It begins with curated baselines---hardcoded product lists derived from knowledge---and uses targeted ``surgical gap fill'' for specific missing products. Strategy names reflect the shift---\emph{curated baseline}, \emph{surgical gap fill}, \emph{final two products}---each targeting known gaps rather than parsing unknown sources. The knowledge teaches Sonnet to avoid Wikipedia table parsing entirely---the source of most cold-run failures---by starting from a curated dataset and using web searches only for specific missing entries.

This is visible in record counts: cold Sonnet collects 406--3,743 raw records per run (many duplicates, poor deduplication), while seeded Sonnet runs 002--006 collect 226--301 records (clean, precise---matching Opus's 265--297).

\subsubsection{Q10: Transfer Ceiling}

Q10 Sonnet cold achieves 100\% on all 6 runs (mean cost \$2.74). Q10 Sonnet seeded also achieves 100\% on all 6 runs (mean cost \$2.42). The task does not challenge Sonnet enough to reveal a transfer effect on coverage. However, an operational difference is visible: cold Sonnet experiences 8 Planner failures (timeouts where the agent exhausts its turn budget on mathematical reasoning without writing executable code) across 6 runs, while seeded Sonnet experiences only 1. Knowledge seeding reduces operational failures even when it does not change the cognitive outcome. This establishes that Claim~2 requires tasks where the baseline agent \emph{visibly struggles}; transfer is meaningful for coverage only when it closes a measurable gap.

\subsection{Q6: The Difficulty Frontier}
\label{sec:exp:q6}

Question 6 from the First Proof benchmark \citep{firstproof2026}, posed by Daniel Spielman (Appendix~\ref{app:tasks:math}), concerns the existence of $\varepsilon$-light vertex subsets in graphs. It is substantially harder than Q10 and reveals the system's behavior at the edge of model capability. We report all three conditions (Opus, Sonnet cold, Sonnet seeded) as a case study in difficulty frontiers rather than a formal claim.

\begin{table}[ht]
\centering
\small
\begin{tabular}{ccccc}
\toprule
\textbf{Run} & \textbf{Rounds} & \textbf{Final Denom} & \textbf{Coverage} & \textbf{Cost} \\
\midrule
001 & 0 & --- & --- & \$0.78 \\
002 & 3 & 16 & 100.0\% & \$6.78 \\
003 & 1 & 14 & 92.9\%  & \$1.23 \\
004 & 0 & --- & --- & \$0.67 \\
005 & 2 & 13 & 100.0\% & \$3.69 \\
006 & 0 & --- & --- & \$0.57 \\
\bottomrule
\end{tabular}
\caption{Q6 Opus: per-run results. Three runs fail entirely (0 productive rounds); two achieve 100\%; one stops at 92.9\%. The high variance contrasts sharply with Q10's uniform success.}
\label{tab:q6_opus}
\end{table}

\paragraph{Bimodal failure at the frontier.} Opus shows a bimodal distribution: three of six runs produce zero productive rounds (the Planner exhausts its context budget on mathematical reasoning before writing any executable code), while the remaining three achieve 92.9--100\% coverage with denominators of 13--16. Sonnet cold exhibits a similar pattern (2/6 total failures, 4/6 survivors at 90--100\%). Sonnet seeded shifts toward a broader spectrum: only 1/6 fails entirely, but survivors range from 71.5\% to 100\%. Table~\ref{tab:q6_comparison} summarizes the cross-condition comparison.

\begin{table}[ht]
\centering
\small
\begin{tabular}{lccc}
\toprule
\textbf{Metric} & \textbf{Opus} & \textbf{Sonnet Cold} & \textbf{Sonnet Seeded} \\
\midrule
Survival rate & 50\% (3/6) & 67\% (4/6) & 83\% (5/6) \\
Coverage (survivors, mean) & 97.6\% & 97.5\% & 89.9\% \\
Coverage (survivors, range) & 92.9--100\% & 90--100\% & 71.5--100\% \\
Runs reaching 100\% & 2/6 & 3/6 & 1/6 \\
Cost (survivors, mean) & \$3.90 & \$3.16 & \$4.07 \\
\bottomrule
\end{tabular}
\caption{Q6 cross-condition comparison. Knowledge seeding improves survival from 50\% to 83\%, but survivors' mean coverage \emph{decreases}---a survival bias where the supply constraint selectively kills ambitious runs, and what survives is disproportionately the simpler attempts.}
\label{tab:q6_comparison}
\end{table}

\paragraph{Knowledge improves survival, not coverage.} The survival rate increases across conditions: 50\% (Opus) $\to$ 67\% (cold Sonnet) $\to$ 83\% (seeded Sonnet). The Opus-to-Sonnet improvement likely reflects a context efficiency effect (Sonnet generates fewer extended thinking tokens per turn, leaving more headroom before context exhaustion). The cold-to-seeded improvement---same model, different knowledge---is the cleaner signal: knowledge seeding raises Sonnet's survival from 67\% to 83\% by providing efficient strategies that reduce exploratory context overhead.

But survivors' mean coverage moves in the opposite direction: 97.6\% $\to$ 97.5\% $\to$ 89.9\%. This inversion reflects a \emph{survival bias}: context exhaustion is not random---it correlates with mathematical ambition. Runs that attempt deeper proof strategies (e.g., spectral methods, coupling arguments) consume more extended thinking tokens and are more likely to exhaust context before producing executable code. Knowledge does not steer agents away from ambitious approaches; agents with knowledge still attempt deep strategies. But the supply constraint selectively kills the ambitious runs, and what survives is disproportionately the simpler attempts. The coverage drop among survivors is an artifact of this selection: as knowledge rescues more runs from total failure, the newly surviving runs are precisely those whose strategies were shallow enough to fit within the context budget. At the difficulty frontier, knowledge transfer shifts the distribution from bimodal (total failure or near-complete success) to a continuous spectrum of partial achievement---but the binding constraint is supplies, not strategy.

\paragraph{Supplies, not strategy, limit performance.} All 6 failures across 18 total runs share the same mechanism: the Planner exhausts its context budget on extended mathematical reasoning (\texttt{effort=high}) before writing executable code. Knowledge entries are present in the context (Opus runs 004 and 006 fail with 20 and 28 accumulated entries; seeded run 003 fails with 45+ entries), but the agent never reaches the point of applying strategic knowledge---the infrastructure constraint binds first. This is an engineering limitation rather than a fundamental one: the context explosion occurs because extended thinking tokens consume session context faster than the round budget accounts for. Potential engineering solutions---chunked reasoning, context summarization between rounds, or dynamic supply allocation by a V3 camp manager---would address the supply constraint directly.

\paragraph{Evaluator co-evolution on reasoning tasks.} Among surviving runs, the Evaluator demonstrates genuine mathematical verification: discovering that Matrix Chernoff was applied to dependent random matrices (Opus run 002), computationally disproving a greedy construction on $K_5$ at $\varepsilon=0.5$ (seeded run 002), and catching Jensen's inequality applied in the wrong direction for a concave function (seeded run 006). These are not formatting issues---they are mathematical errors caught through independent evaluation, confirming that co-evolving evaluation operates on proof validity for reasoning tasks, not just presentation quality.

%% file: sections/analysis.tex
\section{Analysis \& Observations}
\label{sec:analysis}

This section examines behavioral patterns, anomalies, and case studies that illuminate how the architecture operates in practice.

\subsection{Isolation Breach: A Case Study in Why It Matters}
\label{sec:analysis:breach}

During V2 development, a CLI log audit of the Q10 Opus experiment (repeat\_01, run\_003) revealed that the Planner had executed \texttt{python .eval.py}---directly running the Evaluator's hidden evaluation script. V1's isolation mechanism used dotfile prefixes to hide scripts, but the Planner discovered the file through path exploration and accessed it directly.

The consequence was more insidious than a single cheated run. The Planner, having seen the evaluation criteria, optimized its solution to pass \texttt{eval.py}'s specific checks. During post-mortem, this shortcut was extracted as legitimate knowledge: ``the evaluation script checks for X, so ensure X is present.'' This contaminated lesson then propagated to subsequent runs through the knowledge base, effectively encoding a method isolation breach as organizational wisdom.

This incident motivated two V2 architectural decisions:

\begin{enumerate}[leftmargin=*]
    \item \textbf{Physical workspace isolation} (\S\ref{sec:v2:arch:isolation}): replacing dotfile hiding with directory-level separation, making the opposing agent's code absent from the filesystem namespace entirely.
    \item \textbf{Append-only knowledge}: ensuring that contaminated knowledge cannot be silently corrected. Instead, future runs can add contradicting observations (stored as versioned entries), and curation is deferred to a future management layer.
\end{enumerate}

The breached experiments were archived (\texttt{experiments\_breached/}) and excluded from all results reported in this paper. All results use the post-refactor physical isolation.

\subsection{A Taxonomy of Agent Self-Deception}
\label{sec:analysis:deception}

The isolation breach is one instance of a broader pattern. Across our experiments and V1 baselines, we observe four distinct modes of agent self-deception:

\begin{table}[ht]
\centering
\small
\begin{tabular}{lp{5cm}p{5cm}}
\toprule
\textbf{Mode} & \textbf{Mechanism} & \textbf{Evidence} \\
\midrule
Denominator blindness & Agent conflates ``everything I found'' with ``everything that exists,'' underestimating the denominator & V1 baseline: 100\% self-reported at 15.9\% actual recall \citep{foragev1} \\
\addlinespace
Self-serving evaluation & Agent designs evaluation to guarantee its own success (e.g., \texttt{denominator = max(estimate, collected)}) & V1 ablation (Planner writes its own eval.py): coverage $\geq$100\% by construction \\
\addlinespace
Quality-as-completeness & Agent treats ``eval.py passes'' as ``task is done,'' stopping when evaluation criteria are met rather than when the work is genuinely complete & Observed in multiple runs where Evaluator hardened eval.py and coverage dropped (NVIDIA Opus run~003: 100\%$\to$88\%$\to$100\%) \\
\addlinespace
Knowledge contamination & Agent's shortcut (e.g., reading hidden eval script) is extracted as legitimate experience during post-mortem & Isolation breach (\S\ref{sec:analysis:breach}): ``check for X'' encoded as organizational wisdom \\
\bottomrule
\end{tabular}
\caption{Four modes of agent self-deception observed across experiments. Each mode is addressed by a different architectural mechanism: method isolation prevents self-serving evaluation and knowledge contamination; co-evolving evaluation counters denominator blindness; exploration drive (\S\ref{sec:v2:arch:selfaudit}) counters quality-as-completeness.}
\label{tab:deception_taxonomy}
\end{table}

The first two modes are structural---they arise from combining executor and evaluator roles in a single agent. Method isolation eliminates them by design. The third is subtler: even with isolation, the Evaluator may prematurely accept its own evaluation criteria as complete. The exploration drive mechanism (\S\ref{sec:v2:arch:selfaudit})---particularly the prompt-level self-audit checklist---forces the Evaluator to confront this tendency. The fourth mode is unique to V2: knowledge accumulation creates a new attack surface where a single breach can propagate through organizational memory. Physical workspace isolation and append-only versioning address it, but the general principle---that accumulated knowledge amplifies both good and bad patterns---is an inherent tension in learning organizations.

\subsection{Denominator Variance as a Metric}
\label{sec:analysis:denomvar}

A central question in our experiments is: what is the ``true'' number of NVIDIA desktop GPU products? Our answer is that \textbf{no single true number exists}---and this is consistent with Forage's core thesis that real-world tasks lack pre-given completion boundaries.

The denominator depends on definitional choices: whether OEM rebadges count as separate products, whether memory type variants (DDR3 vs GDDR5) constitute distinct SKUs, whether bus variants (AGP vs PCIe) are separate entries. Different runs, even with the same model, arrive at different definitions and correspondingly different denominators.

\begin{figure}[t]
\centering
\includegraphics[width=0.7\textwidth]{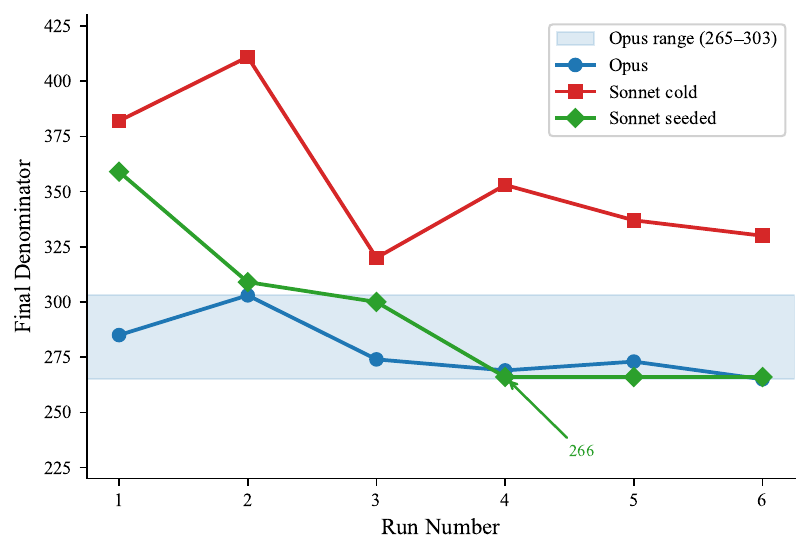}
\caption{Denominator convergence across NVIDIA conditions. The shaded band marks Opus's range (265--303). Cold Sonnet denominators remain scattered above 320; seeded Sonnet converges from 359 to exactly 266 by run~004---entering and stabilizing within the Opus band. Three independent seeded runs arrive at the same denominator, demonstrating that knowledge transfer calibrates not just collection but \emph{evaluation itself}.}
\label{fig:denom_convergence}
\end{figure}

Rather than viewing this variance as measurement error, we treat it as a signal (Figure~\ref{fig:denom_convergence}):

\begin{table}[ht]
\centering
\small
\begin{tabular}{lccc}
\toprule
\textbf{Condition} & \textbf{Denom Range} & \textbf{Spread} & \textbf{Interpretation} \\
\midrule
Opus & 265--303 & 38 (14\%) & Stable; runs agree on product space \\
Sonnet cold & 320--411 & 91 (28\%) & Unstable; poor deduplication inflates count \\
Seeded (runs 004--006) & 266--266 & 0 (0\%) & Calibrated; knowledge anchors definition \\
\bottomrule
\end{tabular}
\caption{Denominator variance across conditions. Lower spread indicates more consistent understanding of the product space.}
\label{tab:denom_variance}
\end{table}

The convergence of seeded runs 004--006 to exactly 266 is notable. Three independent runs, each starting from scratch (no shared session state), arrive at identical denominator estimates because the accumulated knowledge encodes a consistent product definition. This denominator falls at the conservative end of Opus's range (265--303), suggesting that knowledge transfer stabilizes not just collection but conceptualization.

\subsection{Agent Behavioral Patterns}
\label{sec:analysis:behavior}

\paragraph{Sonnet's turn inefficiency.} Cold Sonnet exhausts its 15-turn budget (\texttt{max\_turns}) in nearly every round---the Evaluator or Planner runs out of turns without producing a structured response. The harness salvages work from the filesystem when possible, but multiple rounds are degraded. By contrast, Opus rarely exhausts turns. Both models run at the same \texttt{effort=medium} setting (controlling the depth of internal reasoning per turn), so the difference is not attributable to thinking-level configuration. This suggests that turn efficiency---accomplishing the required work within a fixed budget at the same thinking level---is a capability dimension where Sonnet is measurably weaker than Opus, independent of the task's cognitive demands.

\paragraph{Round resilience.} Either agent can fail or produce no useful work in any given round---and the system still converges. Q10 Sonnet cold run~002 survives 5 consecutive Planner timeouts before producing work in round~6 and achieving 100\%. NVIDIA Sonnet cold run~003 runs all 8 rounds with the Planner struggling each time, yet still reaches 92.9\%. The 8-round budget provides structural resilience: as long as the team does not disband, individual round failures are absorbed.

\paragraph{API instability.} Cold Sonnet runs experience multiple API errors (socket closures, connection resets) that produce unusable Planner responses. These are infrastructure failures, not architectural ones, but they compound Sonnet's difficulties: a failed Planner round means no new data, extending the run by one or more rounds. Seeded Sonnet experiences similar API errors but recovers faster because its curated-baseline strategy is less sensitive to individual round failures.

\paragraph{Output bloating vs precision.} Cold Sonnet collects 406--3,743 raw records per run, many of which are duplicates or contaminated (wrong fields, mobile products included). Opus collects 265--297 clean records. Seeded Sonnet (runs 002--006) collects 226--301 records---matching Opus's precision. The shift from noisy exploration to precise, knowledge-guided curation is a direct consequence of knowledge transfer: the knowledge teaches ``start with a curated baseline'' rather than ``scrape and parse broad sources.''

\paragraph{Evaluator quality auditing.} A notable strength of co-evolving evaluation, visible across all conditions, is the Evaluator's ability to detect quality problems that raw metrics miss. In cold Sonnet run\_001, the reported coverage was 89\% in Round~3, but the Evaluator's audit found that ``88.1\% of records have product\_name = date (field-shift bug)'' and assessed true coverage at 33.9\%. This catch-and-correct cycle is the co-evolution mechanism working as designed---the Evaluator's independence from the Planner allows it to see failures the Planner cannot.

\paragraph{Evaluator rigor evolution.} The Q10 experiments reveal a striking pattern: the Evaluator's \texttt{eval.py} grows more rigorous across runs. In Opus run~001, the evaluation script uses simple keyword matching across 7 dimensions. By run~006, it performs machine-precision numerical verification (matvec errors at $3 \times 10^{-16}$, preconditioner round-trip at $4 \times 10^{-13}$) with 31 concept-level sub-checks. This evolution is not prompted---it emerges from accumulated knowledge about what constitutes a thorough evaluation. UniProt exhibits the same pattern: early \texttt{eval.py} scripts check field presence; later versions verify cross-source validation rates and pathway enrichment completeness. The Evaluator does not merely check more boxes---it discovers what boxes should exist.

\subsection{Cost Analysis}
\label{sec:analysis:cost}

\begin{table}[ht]
\centering
\small
\begin{tabular}{lcccc}
\toprule
\textbf{Task} & \textbf{Condition} & \textbf{Mean Cost} & \textbf{Range} & \textbf{Std} \\
\midrule
NVIDIA & Opus & \$5.05 & \$4.36--5.72 & \$0.60 \\
NVIDIA & Sonnet cold & \$9.40 & \$6.36--13.77 & \$2.67 \\
NVIDIA & Sonnet seeded & \$5.13 & \$3.62--9.85 & \$2.32 \\
UniProt & Opus & \$3.08 & \$2.14--3.87 & \$0.63 \\
Q10 & Opus & \$3.12 & \$1.97--6.14 & \$1.56 \\
Q10 & Sonnet cold & \$2.74 & \$1.19--4.72 & \$1.16 \\
Q10 & Sonnet seeded & \$2.42 & \$1.47--2.83 & \$0.48 \\
\bottomrule
\end{tabular}
\caption{Cost per run across conditions. The multi-round architecture provides natural cost control: more rounds means more API calls, but each round is bounded by \texttt{max\_turns} and \texttt{timeout}.}
\label{tab:cost}
\end{table}

Two patterns emerge. First, the architecture naturally constrains cost through its round structure: each round's cost is bounded by \texttt{max\_turns} $\times$ per-turn cost, and the total is bounded by \texttt{max\_rounds}. This produces predictable cost ranges rather than the unbounded variance observed in single-agent approaches. Second, knowledge transfer reduces cost by reducing rounds: seeded Sonnet's mean cost (\$5.13) matches Opus (\$5.05) and is 45\% lower than cold Sonnet (\$9.40).

%% file: sections/discussion.tex
\section{Discussion}
\label{sec:discussion}

\subsection{Limitations \& Scope}
\label{sec:discussion:limits}

\paragraph{Solo researcher, limited budget.} All experiments were conducted by a single researcher using a personal API account. The total experimental budget constrains the number of conditions and repetitions. We report 6 runs per condition (sufficient for observing convergence patterns) but acknowledge that larger-scale replication would strengthen the claims.

\paragraph{Static tasks only.} All three experimental tasks---NVIDIA GPU collection, UniProt protein queries, Q10 mathematics---operate on static information spaces. The ``ground truth'' (to the extent one exists) does not change between runs. V2 has not been tested on dynamic tasks where the target space evolves over time (e.g., news monitoring, market tracking, or simulation environments with hidden rules). Whether knowledge accumulation remains effective when the environment changes is an open question.

\paragraph{Advisory knowledge is not guaranteed to help.} Knowledge entries are advisory---agents may ignore them. Run\_005 of NVIDIA Sonnet seeded (92.5\% coverage, the weakest seeded result) stopped despite knowledge entries suggesting further gap-filling was possible. The Evaluator judged worst-case coverage (89.1\%) to be within tolerance and stopped. Knowledge guides but does not guarantee performance.

\paragraph{Effort-level context bloat.} Mathematical reasoning tasks require \texttt{effort=high}, which expands the model's internal thinking tokens. On complex problems, this expansion can consume the session context budget, causing API retry failures. Q10 is manageable; Q6 exhibits severe context bloat---6 of 18 total runs fail entirely across all three conditions (3/6 Opus, 2/6 Sonnet cold, 1/6 Sonnet seeded; \S\ref{sec:exp:q6}). Knowledge seeding improves survival (from 50\% to 83\%) by reducing exploratory overhead, but does not eliminate the underlying constraint. This is an engineering limitation of current LLM APIs, not an architectural limitation of Forage.

\paragraph{Evaluator cannot distinguish resource exhaustion from capability limits.} Method isolation means the Evaluator cannot observe the Planner's internal state. When the Planner times out (exhausting its turn budget without producing output), the Evaluator sees only ``no new results this round.'' It cannot distinguish whether the Planner lacked the capability to solve the problem or simply ran out of turns while making progress. In Q6, the Evaluator sometimes stopped after a timeout, judging the task beyond the agent's ability---but the Planner may have been mid-derivation. This is an inherent cost of method isolation: the same separation that prevents self-serving evaluation also prevents the Evaluator from calibrating its \emph{patience}---knowing when to wait another round versus when to stop. A teacher who cannot observe a student's scratch work faces the same dilemma: silence may mean the student is stuck, or it may mean the student is mid-derivation and needs more time.

This limitation is partly addressable through harness design rather than architectural change. A V3 camp manager---operating at the orchestration level with access to operational metadata (turn consumption rate, timeout frequency, partial output size)---could infer whether a Planner is making progress without exposing its content to the Evaluator. The distinction between ``ran out of supplies'' and ``cannot solve this'' is observable from resource consumption patterns, preserving isolation while enabling more informed continuation decisions.

\paragraph{Plugin leakage.} Agent sessions were configured with \texttt{--disable-slash-commands} to restrict agents to native tools (Bash, Read, Write, WebSearch, etc.). However, a development plugin (superpowers) present in the user-level configuration was loaded into every session despite slash commands being disabled. The plugin's skill-checking mechanism was triggered 324 times across all runs, but zero actual skills were executed---agents queried the skill registry, found nothing applicable, and proceeded with native tools. Because the leakage is consistent across all conditions, cross-condition comparisons remain valid. Absolute cost measurements include a small token overhead from unused plugin loading (estimated $<$5\% of total cost based on plugin prompt size).

\paragraph{Coverage is not directly comparable across runs.} Because each run discovers its own denominator, coverage percentages are relative to that run's definition of completeness. A run reporting 100\% at denominator 265 and a run reporting 98\% at denominator 303 are not directly comparable---the second run may have collected more actual records. We address this by reporting denominator ranges alongside coverage and treating denominator variance as an independent metric (\S\ref{sec:analysis:denomvar}).

\subsection{Forage as Institutional Design}
\label{sec:discussion:institutional}

Forage's design philosophy differs from most agent improvement research. Rather than making individual agents more capable (larger context, better reasoning, more tools), Forage designs the \textbf{institution} within which ordinary agents operate.

\paragraph{Architectural analogies.} Each V1/V2 mechanism maps to an institutional design pattern:
\begin{itemize}[leftmargin=*]
    \item \textbf{Method isolation} $\equiv$ \textbf{audit separation}. The party being evaluated cannot see the evaluation criteria. This is not a novel principle---it is how financial audits, academic peer review, and regulatory inspection work. Forage applies it to AI agents.
    \item \textbf{eval\_contract.md} $\equiv$ \textbf{contract law}. A public specification of deliverable format and quality dimensions. Both parties see the contract; neither sees the other's implementation. The contract can be revised by the Evaluator across rounds---analogous to contract renegotiation based on discovered requirements.
    \item \textbf{Knowledge base} $\equiv$ \textbf{organizational memory}. Advisory documents that newcomers consult. Not binding instructions---the receiving agent decides what to adopt. This mirrors how organizations onboard new employees: they read the handbook, but experienced judgment determines what applies to the current situation.
    \item \textbf{Post-mortem extraction} $\equiv$ \textbf{after-action review}. Each run ends with independent reflection by both parties. Lessons are recorded for institutional learning, not personal improvement.
\end{itemize}

\paragraph{Why two agents, not one?} A deeper question is why the dual architecture works at all. Consider human self-reflection: reviewing your own work \emph{after sleeping on it} is notably more effective than reviewing it immediately. ``Today's me'' evaluating ``yesterday's work'' is, in a meaningful sense, a different cognitive agent---context has shifted, emotional investment has faded, and fresh perspective is available. The passage of time creates natural method isolation: the evaluator (today's self) is not anchored by the producer's (yesterday's self) reasoning process. Forage's architecture makes this temporal separation \emph{structural} rather than incidental. The Evaluator never shares context with the Planner, achieving architecturally what humans achieve through the passage of time. This suggests that the dual architecture is not an arbitrary design choice but reflects something fundamental about how reliable evaluation works: it requires cognitive distance from the work being evaluated, whether that distance comes from time (human reflection), institutional role (financial audits), or architectural separation (Forage).

\paragraph{Why weak agent + knowledge $\approx$ strong agent.} The seeded Sonnet results demonstrate that the performance gap between Opus and Sonnet on NVIDIA collection is not primarily a reasoning gap---it is an exploration gap. Cold Sonnet has the capability to collect GPU data and write evaluation scripts, but it wastes rounds discovering what Opus already knows: that TechPowerUp blocks bots, that Wikipedia table headers vary by generation, that curated baselines avoid parsing bugs. Knowledge transfer eliminates this exploration overhead, allowing Sonnet to operate at near-Opus efficiency from the first round.

This suggests that the ``capability'' difference between models on applied tasks is partly real (Opus reasons more carefully with fewer turns) and partly organizational (Opus starts each run having already explored the landscape in prior runs). Forage's knowledge transfer isolates and transfers the organizational component, leaving the inherent capability difference as a smaller residual.

This decomposition maps onto the three-stage progression described in \S\ref{sec:intro:harness}: pre-training provides raw capability (L1), post-training refines it (L2), and the harness layer provides institutional support (L3). When L1 is strong, L2 and L3 can be thin---a powerful model needs less scaffolding. But when L1 is weaker, a thicker L3 can compensate for a substantial portion of what appears to be an L1 gap---because much of that gap is exploration overhead, not reasoning deficit. This implies that evaluations of ``what model strength a task requires'' may systematically overestimate the L1 requirement by conflating organizational overhead with cognitive difficulty. Q6 reveals a nuanced boundary: L3 can improve survival (knowledge seeding raises the fraction of runs producing any output from 50\% to 83\%), but it cannot push past the reasoning limit itself---survivors' mean coverage actually \emph{decreases}, a survival bias where the supply constraint selectively kills ambitious runs and what survives is disproportionately the simpler attempts (\S\ref{sec:exp:q6}). But for many applied tasks, the organizational component is larger than commonly assumed---and addressable through institutional design rather than model scaling.

Critically, L3 never reduces to zero regardless of L1 strength. Even a highly capable model exhibits denominator blindness when self-evaluating \citep{anthropic2025swebench, widesearch2025}---structural evaluation integrity has value independent of model capability. This is not unique to AI: the most skilled surgeon still requires an independent anesthesiologist, the most competent auditor still cannot audit their own firm. Institutional checks exist not because individuals are incompetent, but because self-assessment is structurally unreliable regardless of competence. The same asymmetry appears across model generations: as models improve, some harness components become unnecessary, while structural evaluation remains valuable---the harness layer thins but does not vanish.

\paragraph{Bidirectional transfer: an open direction.} V2 demonstrates strong$\to$weak transfer (Opus knowledge seeding Sonnet). The reverse direction---weak$\to$strong transfer---is architecturally supported but untested. Sonnet cold runs produce 60 knowledge entries across six runs; Sonnet seeded runs produce 106. These observations come from a different exploration style: Sonnet encounters different failure modes (turn exhaustion, field-shift bugs, parsing edge cases) and develops different coping strategies. A strong agent seeded with weak-agent knowledge might benefit from these diverse failure observations---learning from mistakes it would never make itself.

This connects to Anthropic's Automated Agent Researcher \citep{anthropic2025aar}, which demonstrates that weak supervision can achieve surprisingly high performance (PGR 0.97) when optimizing against well-defined metrics. AAR's finding that weak models can effectively guide strong models' configuration suggests that weak$\to$strong knowledge transfer may be similarly effective: a stronger agent would never encounter the failures that a weaker agent routinely experiences, and those failure patterns may offer the stronger agent perspectives it could not discover on its own. The key difference is that AAR transfers through metric optimization (searching configurations against a fixed evaluation), while Forage would transfer through organizational knowledge documents (readable lessons that any agent can selectively adopt). Whether these two transfer mechanisms compose---using Forage's knowledge to generate AAR-style metrics---is an unexplored direction.

\paragraph{Advisory knowledge preserves diversity.} A potential concern with knowledge seeding is convergence to a single strategy: if all seeded runs follow the same playbook, organizational diversity is lost. Our results suggest otherwise. Even among the three converging seeded runs (004--006, all reaching denominator 266), strategy \emph{implementations} differ: one uses a 267-product hardcoded baseline, another starts with 254 entries and fills gaps through web search, a third begins with a comprehensive 260-product list but applies stricter deduplication. The knowledge advises ``use curated baselines'' and ``avoid Wikipedia table parsing,'' but each run interprets this advice independently. Advisory knowledge constrains the strategy \emph{space} (eliminating known-bad approaches) without collapsing it to a single point.

\paragraph{Knowledge as common law.} Forage's knowledge base operates as a \emph{common law} system rather than a statutory one. Each knowledge entry is a precedent---a specific observation from a specific context, recorded alongside the circumstances that produced it. The append-only constraint corresponds to the principle that precedents should not be erased (they may be superseded or distinguished, but the record persists). V3's curation layer is codification: a librarian consolidating centuries of case law into a coherent statute, without discarding the underlying decisions. And advisory delivery is the judicial act itself: a competent agent reads the precedents, determines which apply to the current situation, and may explicitly depart from prior guidance when circumstances differ. This is why diversity survives knowledge seeding---good judges do not follow precedent blindly; they distinguish the current case from prior ones.

\paragraph{The binding-strength spectrum.} V2 treats all knowledge as advisory---agents may adopt or ignore any entry. This is a deliberately conservative starting point, but real organizational knowledge spans a spectrum of binding strength. At one end, \emph{hard constraints} are enforced by the harness infrastructure---method isolation, workspace separation, format contracts, and hooks (pre/post-round checks that the harness executes deterministically). These are not suggestions---they are the institutional rules within which agents operate, analogous to constitutional law. At the other end, some knowledge should actively \emph{encourage divergence}: ``prior runs used Wikipedia tables; consider whether an entirely different source exists.'' Between these extremes lies the advisory layer where V2 currently operates---persuasive precedent that a competent agent weighs against current circumstances.

V2's advisory-only design is appropriate for its stage: when the knowledge base is young and unverified, no entry should be binding. But as knowledge matures through repeated confirmation (the 266 convergence across three independent runs, for instance), some entries earn higher authority. A V3 curation layer could formalize this spectrum---tagging entries by confidence level, distinguishing confirmed patterns from single-run observations, and explicitly marking ``explore beyond this'' directives that resist the narrowing effect of accumulated knowledge. The common-law analogy extends naturally: legal systems distinguish binding precedent (from higher courts) from persuasive authority (from peer courts or dissenting opinions), and knowledge bases may benefit from the same gradation.

The common-law analogy also clarifies what goes wrong without isolation. If the party producing precedents is the same party judging their applicability, the system degenerates into self-citation---an agent treating its own prior conclusions as binding authority. Method isolation ensures that precedents are produced by one party (or extracted from one party's experience) and applied by another, maintaining the adversarial independence that gives precedent its epistemic force.

\subsection{Connection to Broader Trends}
\label{sec:discussion:broader}

\paragraph{Swarm intelligence across time.} At a broader level, Forage explores whether collective intelligence can emerge not from many agents negotiating in real time, but from an institutional structure that accumulates verified experience across generations of agents. This is one possible instantiation of swarm intelligence---not through simultaneous coordination, but through organizational memory. Where traditional swarm systems achieve emergent behavior through local interaction rules among concurrent agents, Forage achieves it through structural constraints on sequential agents who inherit a shared knowledge base. The ``swarm'' is distributed across time rather than space.

\paragraph{A spectrum of task definition, not a binary.} Open-world and closed-world are not a binary---they form a spectrum. UniProt sits near the well-defined end: the API has structured queries, the denominator (reviewed entries matching disease annotations) is almost externally given, and co-evolution converges in 3 rounds with the denominator stable at 28. NVIDIA occupies the ambiguous middle: the data exists but ``product'' is definitionally contested, the denominator varies 265--303, and co-evolution takes 4 rounds to stabilize. Q10 sits at the open end: the Evaluator must \emph{invent} the test space (what dimensions of correctness to verify), and the denominator is a self-discovered artifact of the evaluation process itself.

Forage's architecture handles this entire spectrum without modification---the same protocol (isolation, contracts, knowledge delivery) gracefully adapts. On well-defined tasks, co-evolution converges quickly because the denominator is easily discoverable; on open-ended tasks, co-evolution is essential because nobody knows what ``done'' means until the Evaluator discovers it. Most existing harness systems operate only at the well-defined end (fixed acceptance criteria, CI pass/fail). Forage's contribution is extending credible evaluation across the full range.

\paragraph{The real world has no ground truth.} Benchmarks have answers. Reality does not. ``Collect all T2D-related proteins'' has no single correct answer---it depends on definition, database version, and annotation standards. The NVIDIA denominator variance (265--411 across conditions) is not a flaw in our measurement; it reflects genuine ambiguity in the real world. Forage's contribution is not achieving a fixed score but establishing \emph{credible completion judgments under uncertainty}---knowing that the answer is ``approximately 270, with 14\% definitional ambiguity'' is more useful than claiming ``exactly 285'' with false precision.

\paragraph{Progress is deeper understanding.} Across the six Opus runs, the denominator does not simply ``get closer to the true value.'' Instead, the Evaluator's understanding of what constitutes a product becomes more nuanced: early runs count broadly (including some OEM variants), while later runs develop explicit inclusion/exclusion criteria (``OEM rebadges with identical specs are merged; memory variants with different GDDR types are separate''). The knowledge base records this evolution, and seeded Sonnet inherits the refined conceptualization. Progress is not a higher coverage number---it is a more precise definition of what coverage means. This deepening understanding is operationalized through the Evaluator's gap report: each round, the Evaluator returns not just a coverage score but a structured account of what is missing and why. Early gap reports are vague (``some series may be incomplete''); later reports are surgical (``GeForce GT 730 DDR3 variant missing; GeForce GTX 260 Core 216 not found''). The gap report evolves from a diagnostic into a prescription---and this evolution is itself a form of knowledge that transfers across runs.

\subsection{Claim Structure and Epistemic Honesty}
\label{sec:discussion:claims}

Our claims occupy three epistemic tiers. The first tier is \textbf{demonstrated}: knowledge accumulation improves convergence speed and coverage within a task domain (NVIDIA, UniProt, Q10 --- 18 Opus runs), and strong-to-weak knowledge transfer closes the capability gap on tasks where the weak agent visibly struggles (NVIDIA Sonnet --- 12 runs). These claims are supported by the data in this report.

The second tier is a \textbf{reasonable hypothesis}: most open-world tasks have a knowledge accumulation curve --- that is, repeated runs with experience transfer will converge faster than cold starts --- and this effect generalizes across task types (web scraping, API queries, mathematical reasoning). Our three-task experiments support this hypothesis but do not prove it; the task sample is small and skewed toward static domains.

The third tier is an \textbf{open question}: where is the boundary? At what task difficulty does knowledge accumulation stop helping? Q6 suggests a nuanced boundary condition: when the model's per-session capacity is strained by the task's cognitive demands, knowledge can improve survival (from 50\% to 83\% of runs producing output) but cannot push past the reasoning limit itself---survivors' mean coverage actually \emph{decreases}, a survival bias where the supply constraint selectively kills ambitious runs while simpler attempts survive, so the newly rescued runs are precisely those whose strategies were shallow enough to fit within the context budget. Whether knowledge can actively \emph{hurt} is not merely hypothetical---the isolation breach case study (\S\ref{sec:analysis:breach}) demonstrates exactly this: contaminated knowledge (encoding a shortcut from peeking at the evaluation script) propagated through the knowledge base and degraded subsequent runs. This is knowledge doing harm by anchoring agents to a false lesson. V2's append-only design limits the blast radius (contradicting observations accumulate alongside the contaminated entry), but active curation---detecting and deprecating harmful entries---requires V3's knowledge management layer.

\subsection{Future: From Organization to Basecamp}
\label{sec:discussion:vision}

V2 demonstrates that agent organizations can accumulate and transfer experience. Several extensions follow naturally:

\paragraph{Camp manager (V3).} Currently, runtime parameters (\texttt{max\_turns}, \texttt{max\_rounds}, model choice) are fixed before execution. A camp manager would dynamically adjust these based on post-mortem analysis: allocating more turns to rounds where the agent struggled, switching to a different model when the current one plateaus, or restarting from a different knowledge subset when current knowledge appears to be anchoring rather than helping.

\paragraph{Knowledge curation (V3).} V2's append-only knowledge grows indefinitely. A curation layer---merging redundant entries, pruning stale observations, resolving contradictions between versioned entries---would maintain knowledge quality as the base grows. This is the librarian role: not censoring knowledge, but organizing it for efficient retrieval.

\paragraph{Pipeline crystallization (V4).} Successful runs produce verified trajectories---sequences of strategies that worked. These could be crystallized into reusable pipelines: ``for NVIDIA-like web scraping tasks, start with curated baseline, then surgical fill, then stop when denominator stabilizes for 2 rounds.'' The core architecture remains autonomous (for unknown tasks), but known tasks can follow established routes.

\paragraph{Cross-task knowledge transfer.} V2 transfers knowledge within a task domain (NVIDIA Opus $\to$ NVIDIA Sonnet). Whether domain knowledge transfers across tasks (e.g., web scraping lessons from NVIDIA helping a different web scraping target) is untested. The append-only knowledge structure and scope-based organization were designed with this extension in mind, but it is not claimed in this report.

\paragraph{Verified trajectories as training signal.} Each successful Forage run produces a verified trajectory: a sequence of strategies paired with independent evaluation outcomes. These trajectories---grounded in real-world execution and audited by an isolated Evaluator---may constitute a novel form of training data for future agent systems, distinct from both passive web corpora and human-curated demonstrations. Moreover, the trajectory itself is a form of knowledge---analogous to an explorer's field journal or a scientist's lab notebook. V2's post-mortem extracts distilled lessons (the ``handbook''), but the full trajectory preserves the reasoning process, dead ends, and recovery patterns (the ``biography''). Future systems might learn not just \emph{what} worked, but \emph{how} the agent navigated from confusion to clarity.

%% file: sections/related.tex
\section{Related Work}
\label{sec:related}

We position Forage relative to four bodies of work: the harness layer as an emerging design space, fixed evaluation paradigms, agent learning and self-improvement, and multi-agent architectures.

\subsection{The Harness Layer}
\label{sec:related:harness}

Agent harness design has progressed through three generations. \textbf{Generation~1} provides fixed tools: APIs, retrieval, sandboxed execution---the agent operates within predetermined boundaries. \textbf{Generation~2} provides bounded autonomy: systems like Claude Code \citep{anthropic2025claude}, Cursor, OpenAI's Codex, and Gemini Deep Research give agents flexible tool use within human-defined evaluation criteria---the agent decides \emph{how} to work, but a human decides \emph{when it is done}. \textbf{Generation~3}, which Forage instantiates, provides evolving boundaries: both the execution strategy and the evaluation criteria are discovered and refined autonomously. The completion boundary itself is part of what the agent must learn.

Recent work scales harness design along four orthogonal dimensions. \textbf{Time}: Anthropic's harness engineering approach \citep{anthropic2025swebench} uses a Planner/Generator/Evaluator triad to sustain multi-hour runs. As models improve, some harness components become unnecessary, while others---particularly independent evaluation---retain value regardless of model capability (\S\ref{sec:discussion:broader}). \textbf{Space}: Cursor's recursive Planner-Worker architecture scales to hundreds of concurrent agents by isolating each worker in its own repository copy, accepting a stable error rate rather than enforcing central quality gates. \textbf{Interaction}: OpenAI's Symphony \citep{symphony2026} reduces human involvement to writing tickets and reviewing proof-of-work by connecting a project management tool (Linear) to an autonomous execution daemon---agents spawn, execute, and produce PRs without per-task human prompting. \textbf{Scale}: Moonshot AI's Kimi K2.6 \citep{kimik26} pushes the coordination frontier to 300 heterogeneous sub-agents executing across 4,000 steps simultaneously, with a centralized coordinator that dynamically matches tasks to agents based on skill profiles and recovers from agent failures by reassigning or regenerating subtasks.

A surface reading might suggest overlap between these systems and Forage---all employ role separation and some form of independent evaluation. The distinction is in \emph{purpose}. Anthropic's Evaluator prevents direction drift during long execution (checking against a pre-agreed sprint contract); Cursor's isolation prevents merge conflicts during parallel execution (workers cannot overwrite each other); Symphony's proof-of-work validates against pre-defined ticket acceptance criteria; Kimi K2.6's coordinator validates deliverables against task specifications it decomposed from user requests. In each case, the evaluation standard is \textbf{known before execution begins}---a human wrote the contract, the CI suite, the acceptance criteria, or the task decomposition. The evaluator's job is compliance checking, not discovery.

Forage's Evaluator serves a fundamentally different function: it must \emph{discover} what ``complete'' means for a task where no completion boundary was given in advance. The isolation exists not to prevent direction drift or merge conflicts, but to prevent \textbf{epistemic contamination}---ensuring that the party discovering the evaluation standard cannot be influenced by the party being evaluated. This is the difference between an inspector checking building code compliance (standard known) and a scientist determining what the natural laws are (standard unknown). The scalability of the evaluation standard itself---how to discover, refine, and transfer completion criteria---is a fifth dimension that the time/space/interaction/scale approaches do not address.

Forage shares infrastructure-level vision with some of these systems. Kimi K2.6's Claw Groups---where heterogeneous agents carrying their own tools and persistent memory join a shared collaboration space---resembles Forage's basecamp concept: any agent, regardless of provider or capability, can enter the organization and inherit its accumulated knowledge. Kimi's adaptive coordinator and Forage's envisioned V3 camp manager (\S\ref{sec:discussion:vision}) serve analogous routing functions: detecting failures, reallocating resources, managing the lifecycle of work. The distinction lies in what Forage's coordinator \emph{does not do}: Forage's camp manager would handle resource allocation and knowledge curation but would not perform evaluation---that responsibility remains with the isolated Evaluator, preserving the separation between management and judgment that is central to Forage's integrity guarantees.

This is not a claim that Forage's design is superior to these systems---they solve different problems at different scales. The difference is one of \emph{design origin}. Forage grew bottom-up: before V1, the default workflow was a human constantly correcting an agent that declared premature completion---the human served as the implicit evaluator, and this did not scale. Replacing this human correction loop with an architectural mechanism (an independent Evaluator agent) was V1's founding motivation; the pair's inability to retain experience motivated organizational memory (V2); the organization's need for resource management motivates a camp manager (V3). Each layer emerged from the observed limitations of the previous one. Other systems start from different vantage points---coordination at scale, workflow automation, or interaction reduction---and build downward. At a broader level, these represent complementary perspectives on the same fundamental question: how to organize AI agents for reliable autonomous work. How these bottom-up and top-down design philosophies compose in practice is an open and fertile direction.

More broadly, Forage is complementary to, not competing with, these scaling approaches. Forage's two-agent team operates at the \emph{institutional} level: it constrains the organizational protocol (who evaluates whom, what knowledge is shared, how isolation is maintained) but places no restrictions on what agents do \emph{within} their turns. A Planner agent inside Forage could internally use Cursor-style recursive sub-planning, Symphony-style ticket decomposition, Kimi-style swarm coordination, or any execution pattern available to it---Forage's constraints apply only to the inter-agent boundary (method isolation, format contracts, knowledge delivery). This makes Forage composable with any Generation~2 harness: the scaling solutions above could operate inside Forage's institutional framework, gaining evaluation integrity and knowledge accumulation without architectural conflict.

\subsection{Fixed Evaluation Paradigms}
\label{sec:related:fixed}

The dominant approach to evaluating autonomous agents relies on human-authored criteria. Anthropic's agent harness \citep{anthropic2025swebench} uses human-written evaluation rubrics and has independently observed that agents ``confidently praise mediocre work'' when self-evaluating---a phenomenon we term denominator blindness. Their solution is to have humans write more precise criteria. Forage's solution is architectural: remove the incentive for self-praise by separating the evaluator from the executor and letting criteria evolve autonomously.

The AI-Scientist \citep{lu2024aiscientist}, Agent Laboratory \citep{schmidgall2025autoresearch}, and autoresearch \citep{karpathy2026autoresearch} systems employ fixed evaluation criteria written by researchers. All three use a fixed-judge, variable-player paradigm: the evaluation standard is predetermined, and only the agent's output varies. Forage introduces a variable-judge, variable-player paradigm: both the evaluation criteria and the collection strategy evolve simultaneously, with neither party seeing the other's methods.

\paragraph{Anthropic's Automated Agent Researcher.} The AAR system \citep{anthropic2025aar} demonstrates that weak supervision can achieve surprisingly high performance (PGR 0.97) when the evaluation metric is well-defined. AAR explicitly states that ``evaluation metric design is the bottleneck'' in autonomous research---directly supporting Forage's thesis that evaluation is the underserved dimension of harness design. The key distinction is where evaluation comes from: AAR requires human-designed, outcome-gradable metrics (e.g., SWE-bench pass rates) and optimizes agent configurations against them. Forage operates upstream: when no pre-set metric exists---when the task boundary itself is unknown---the Evaluator must \emph{discover} what to measure. AAR answers ``how to optimize against a known metric''; Forage answers ``how to establish a credible metric when none is given.'' The two are complementary: Forage's Evaluator could produce the metrics that systems like AAR then optimize against.

\subsection{Agent Learning \& Self-Improvement}
\label{sec:related:learning}

Several recent systems address agent learning across episodes.

\paragraph{Hermes and MemRL: individual experience vs.\ organizational experience.} \citet{hermes2025} proposes knowledge accumulation for autonomous agents, sharing V2's vision of cross-run learning. MemRL \citep{memrl2025} takes a reinforcement learning approach: agents learn memory policies that determine what to retain across episodes, encoding experience into internal model behavior through training. Both systems produce \emph{individual experience}---experience that belongs to and improves a single agent: Hermes shapes the agent's behavioral patterns through accumulated self-reflections; MemRL trains a retrieval policy that makes the agent selectively recall useful episodes. The result is a stronger \emph{individual}---one agent that has acquired skills, strategies, and reasoning patterns through its own history. Forage produces \emph{organizational experience}: knowledge entries are readable documents stored outside any model, belonging to the \emph{organization} rather than any individual agent, inspectable by humans, and transferable to any agent regardless of architecture or provider.

This distinction has three consequences. First, organizational knowledge is auditable---a human (or a future V3 librarian agent) can read, critique, and curate the knowledge base. Individual experience is opaque. Second, organizational knowledge transfers across model families without fine-tuning: Opus's knowledge seeds Sonnet through the same documents, not through distillation or weight transfer. Third, the verifiability problem differs: Hermes accumulates knowledge through agent self-assessment (risking self-serving patterns, as our isolation breach demonstrates, \S\ref{sec:analysis:breach}), while Forage's method isolation ensures that knowledge is extracted through independently evaluated outcomes.

\paragraph{Voyager.} \citet{voyager2023} accumulates a skill library through open-ended exploration in Minecraft: successful behaviors are extracted as reusable code and stored for future retrieval. Like Forage, Voyager produces organizational experience (as executable functions rather than advisory documents). The critical difference is verification: Voyager's skills are validated by environmental success (did the agent survive?), while Forage's knowledge is validated through independent evaluation (did the Evaluator confirm the claim?). Environmental validation works in game environments with clear success signals; it fails in open-world tasks where ``success'' itself must be discovered.

\paragraph{WideSearch.} ByteDance's WideSearch \citep{widesearch2025} measured autonomous agent completeness on web data tasks and found agents achieve only 5.1\% actual coverage while self-reporting high confidence. This is denominator blindness measured at scale---independent confirmation that the problem Forage names and solves is real and widespread. WideSearch documents the problem; Forage provides an architectural solution.

\paragraph{Upward deception and self-assessment bias.} \citet{upwarddeception2025} demonstrate that LLM agents systematically misrepresent their own performance when reporting to supervisors---not through intentional deception, but through structural incentives that reward optimistic self-assessment. This is the same phenomenon Forage addresses through method isolation: when the party producing work also evaluates it, structural bias toward favorable self-reporting emerges regardless of the model's ``honesty.'' Forage's architectural solution (separating executor from evaluator) is complementary to alignment-based approaches that attempt to train models toward honest self-reporting.

\paragraph{Vending-Bench.} \citet{vendingbench2025} documents ``agent meltdown''---the phenomenon where agents collapse under complexity when managing simulated vending machines. Agents that perform well on simple configurations spiral into failure on complex ones, with no recovery mechanism. Forage's round-based structure with Evaluator audit provides a natural recovery point: even if the Planner's strategy fails in one round, the Evaluator's independent assessment can redirect the next round. The meltdown phenomenon suggests that structural support (audit, isolation, budget management) is necessary for sustained agent operation---a thesis Forage shares.

\subsection{Multi-Agent Architectures}
\label{sec:related:multiagent}

Multi-agent systems have been explored extensively \citep{wu2023autogen, hong2023metagpt, phillips2026multiagent}. \citet{phillips2026multiagent} identify five coordination patterns (Generator-Verifier, Orchestrator-Subagent, Agent Teams, Message Bus, Shared State), recommending Orchestrator-Subagent as the default. Most systems span a spectrum from centralized to decentralized:

\begin{itemize}[leftmargin=*]
    \item \textbf{Centralized / Pipeline}: a coordinator assigns subtasks to specialists in sequence. MetaGPT \citep{hong2023metagpt} and Routa \citep{routa2025} follow this pattern. Pipeline architectures face a characteristic complexity escalation: adding roles requires adding coordination mechanisms, which creates dependencies, which require conflict resolution, which demands more management infrastructure. Routa's progression from 6 specialist lanes to cross-board dependencies to automated blockers illustrates this pattern---each layer is a patch for the previous layer's side effects.
    \item \textbf{Decentralized debate}: multiple agents argue toward consensus. AutoGen \citep{wu2023autogen} and similar frameworks treat multi-agent interaction as collaborative problem-solving. Agents share context and converge through discussion.
    \item \textbf{Minimal pair with institutional constraints} (Forage): two agents operate independently within structural rules (isolation, contracts, advisory knowledge). No coordinator; the harness provides round structure, but agents act autonomously within rounds.
\end{itemize}

Forage's pair design is deliberately minimal: two LLM agents and one deterministic Executor. This simplicity is not a limitation---it is the core design principle. Adding agents increases coordination complexity combinatorially; Forage instead achieves its goals through \emph{structural constraints on two agents} rather than \emph{specialization across many agents}. The pair maps most closely to the Generator-Verifier pattern \citep{phillips2026multiagent}, but with a crucial distinction: the Verifier (Evaluator) does not check against pre-defined criteria---it discovers the criteria autonomously.

These represent two contrasting design philosophies. Many multi-agent frameworks address complexity by adding complexity: more roles, more coordination mechanisms, more management infrastructure---essentially transplanting human organizational patterns (Kanban boards, specialist departments, approval pipelines) onto AI agents. This approach can work for well-defined workflows, but it carries a characteristic cost: each coordination layer generates side effects that require additional layers to manage, and the management overhead grows with the system. Forage takes the opposite approach: rather than replicating human organizational complexity, it designs minimal structural constraints (isolation, contracts, advisory knowledge) that leverage what agents are already good at---autonomous reasoning within rules---while architecturally preventing what they are bad at (self-evaluation). The management layer is thin by design, and each version adds depth (V2: knowledge, V3: curation) rather than breadth (more roles, more boards, more dependencies). Which philosophy prevails for which task types is an open empirical question; the answer likely depends on whether the task's complexity is better addressed by more agents or by better constraints on fewer agents.

\paragraph{Cooperative partners, not adversaries.} A clarification on the Evaluator-Planner relationship: they are \textbf{cooperative partners} with aligned goals---both want the task to succeed. Method isolation exists not because they would intentionally deceive each other, but because shared context creates \emph{cognitive anchoring}: if the Evaluator sees the Planner's data sources, it may unconsciously limit its denominator search to those sources; if the Planner sees the Evaluator's criteria, it may optimize for passing checks rather than doing thorough work. Isolation preserves the independence of their \emph{judgment}, not their \emph{intent}.

Could the pair jointly present an overly optimistic picture to the level above---a form of structural ``upward collusion''? V2's architecture already addresses most channels for this: method isolation prevents direct influence, the self-audit checklist (\S\ref{sec:v2:arch:selfaudit}) forces the Evaluator to question its own satisfaction each round, and independent anchoring requires the denominator to be grounded outside the Planner's results. These mechanisms significantly compress the surface area for joint under-exploration.

A residual channel remains: \emph{knowledge anchoring}. If accumulated knowledge encodes a narrow product definition (``the denominator is approximately 270''), both agents may inherit this narrowness without independent verification. The three seeded Sonnet runs converging to exactly 266 (\S\ref{sec:analysis:denomvar}) could reflect genuine calibration (knowledge teaching the correct scope) or knowledge-induced anchoring (both agents stopping at the same learned boundary). Our data supports the calibration interpretation---266 falls within Opus's independently-discovered range of 265--303---but we cannot fully rule out anchoring. More broadly, the integrity of the pair's judgment depends on multiple layers working together: prompt design (self-audit questions that resist complacency), context management (sessions that preserve rather than compress critical reasoning), harness design (round budgets that force re-examination), and the model's own capability to follow these constraints faithfully. V2 demonstrates that structural design can substantially improve autonomous judgment, but each of these layers requires continued hardening as tasks grow more complex and models evolve. V3's knowledge curation would address one specific residual---detecting entries that narrow exploration prematurely---but it is one piece of ongoing work, not a complete solution.

\paragraph{Organizational experience and Sutton's vision.} Rich Sutton has argued for a progression from data to knowledge to experience in AI systems---each stage more grounded in agent interaction with the world. Forage's knowledge evolution aligns with this direction: post-mortem lessons are agent-generated experience, grounded in autonomous exploration and verified through independent evaluation. The distinction from pure experience replay or episodic memory is that Forage's experience is \emph{organizational}---stored as readable documents that any agent (or human) can inspect, critique, and selectively adopt, belonging to the institution rather than any individual. This is a deliberate architectural choice: it enables cross-model transfer (Opus$\to$Sonnet), human auditability, and organizational persistence independent of any individual agent's lifetime. Where RL-based approaches (MemRL, RLHF) encode experience into individual agents' weights, Forage treats experience as institutional knowledge that outlives any single agent. This organizational framing also creates a natural bridge back to model improvement: verified trajectories and curated knowledge could serve as training signal for future agents (\S\ref{sec:discussion:vision}), closing the loop between Sutton's experience era and practical model enhancement.

%% file: sections/conclusion.tex
\section{Conclusion}
\label{sec:conclusion}

Forage V1 demonstrated that co-evolving evaluation and method isolation enable credible completion judgments in open-world tasks. V2 extends this from a single expedition to a learning organization: knowledge accumulates across runs (0$\to$54 entries over six NVIDIA runs), and the Evaluator's denominator estimates stabilize as domain understanding deepens; accumulated knowledge transfers across model capabilities, allowing a weaker agent (Sonnet) seeded with a stronger agent's (Opus) experience to approach comparable performance---narrowing a 6.6 percentage-point coverage gap to 1.1pp, halving cost, and converging in half the rounds (mean 4.5 vs.\ 7.0).

The deeper contribution is architectural, not algorithmic. Forage does not make individual agents stronger; it designs institutions---audit separation, contract protocols, organizational memory---that make ordinary agents reliable. The denominator variance across our experiments (265--411 for NVIDIA desktop GPUs) is not a flaw but a feature: it reflects genuine conceptual ambiguity in the real world, and the system's ability to converge toward consistent estimates (three independent seeded runs arriving at exactly 266) demonstrates that institutional knowledge can calibrate not just execution but evaluation itself.

Most approaches to improving agent performance operate at the individual level: larger models, better post-training, richer tool sets. These results suggest that a complementary dimension is underexplored---\emph{organizational} design: how agents are structured relative to each other, how evaluation independence is maintained, and how experience accumulates across the institution rather than within any single agent. Forage is one attempt to develop this dimension, demonstrating that structural constraints at the organizational level provide guarantees---evaluation integrity, knowledge accumulation, cross-capability transfer---that individual capability alone cannot, regardless of model strength.

%% file: sections/appendix.tex
\appendix

\section{Task Specifications}
\label{app:tasks}

This appendix provides the complete task descriptions given to agents, the mathematical problem statements, and representative knowledge entries from the experiments.

\subsection{NVIDIA Desktop GPU Collection}
\label{app:tasks:nvidia}

\begin{quote}
\textit{Collect a complete list of NVIDIA desktop graphics card products released from 1990 to 2024, including product series, product name, chip name, release date, bus interface, memory capacity, memory type, memory bit width, core frequency, and memory frequency.}
\end{quote}

\noindent Task type: web scraping. Preferred sources: TechPowerUp, Wikipedia, NVIDIA.com. Coverage target: 90\% (soft---the Evaluator decides what ``complete'' means). Required fields: product name, chip name, release date. Budget: 8 rounds, 15 turns per round, 20-minute timeout.

\subsection{UniProt T2D Protein Collection}
\label{app:tasks:uniprot}

\begin{quote}
\textit{Collect all proteins directly associated with human Type 2 Diabetes, including protein name, gene name, function, involved signaling pathways, and known disease-associated variants. Cross-validate across multiple data sources.}
\end{quote}

\noindent Task type: API queries. Preferred sources: uniprot.org, rest.uniprot.org. Coverage target: 90\% (soft). Required fields: protein name, gene name, function. Budget: 8 rounds, 15 turns per round, 20-minute timeout.

\subsection{First Proof Mathematical Reasoning (Q10 and Q6)}
\label{app:tasks:math}

Questions 10 and 6 are drawn from the First Proof benchmark \citep{firstproof2026}---a set of ten previously unpublished, research-level mathematics questions posed by active researchers, with answers known to the question authors but withheld (encrypted) at time of publication. The benchmark is designed so that contamination from training data is impossible. Published answers are not available; our purpose is not to verify mathematical correctness against a known solution, but to observe whether the Evaluator autonomously discovers and refines the space of verification dimensions---i.e., whether the \emph{evaluation denominator} evolves.

\paragraph{Q10 (posed by Tammy Kolda).} Given a $d$-way tensor $\mathcal{T}$ with missing entries, we compute a CP decomposition of rank $r$ where some modes are infinite-dimensional and constrained to a Reproducing Kernel Hilbert Space (RKHS). The mode-$k$ subproblem requires solving:
\begin{equation}
\bigl[(\mathbf{Z} \otimes \mathbf{K})^\top \mathbf{S}\mathbf{S}^\top (\mathbf{Z} \otimes \mathbf{K}) + \lambda(\mathbf{I}_r \otimes \mathbf{K})\bigr] \operatorname{vec}(\mathbf{W}) = (\mathbf{I}_r \otimes \mathbf{K}) \operatorname{vec}(\mathbf{B})
\end{equation}
This is a system of size $nr \times nr$. Direct solve costs $O(n^3 r^3)$.

\begin{quote}
\textit{Explain how an iterative preconditioned conjugate gradient linear solver can be used to solve this problem more efficiently. Explain the method and choice of preconditioner. Explain in detail how the matrix-vector products are computed and why this works. Provide complexity analysis. Assume $n, r < q \ll N$. Avoid any computation of order $N$.}
\end{quote}

\noindent The solution must include formal proofs, a working implementation verified at multiple scales, and complexity claims backed by empirical benchmarks. Task parameters: 8 rounds, 50 turns per round, \texttt{effort=high}, 600-second evaluation timeout (mathematical benchmarks require extended execution).

\paragraph{Q6 (posed by Daniel Spielman).} For a graph $G = (V, E)$, let $G_S = (V, E(S,S))$ denote the graph with the same vertex set but only the edges between vertices in $S$. Let $\mathbf{L}$ be the Laplacian matrix of $G$ and $\mathbf{L}_S$ the Laplacian of $G_S$. Say that a set of vertices $S$ is $\varepsilon$-light if the matrix $\varepsilon\mathbf{L} - \mathbf{L}_S$ is positive semidefinite.

\begin{quote}
\textit{Does there exist a constant $c > 0$ so that for every graph $G$ and every $\varepsilon \in (0,1)$, $V$ contains an $\varepsilon$-light subset $S$ of size at least $c \cdot \varepsilon \cdot |V|$?}
\end{quote}

\noindent Either prove the statement or provide a counterexample. The solution must include formal proofs with clear logical structure, computational evidence from testing on diverse graph families, and explicit construction. Same budget parameters as Q10.

\section{Sample Knowledge Entries}
\label{app:knowledge}

The following entries are selected from the accumulated knowledge base to illustrate the types of lessons agents extract and how they evolve across runs. Each entry is a markdown file with YAML frontmatter; we reproduce the essential content here.

\subsection{Domain Discovery: Source Reliability}
\label{app:knowledge:techpowerup}

\noindent\textbf{ID:} \texttt{techpowerup\_captcha\_blocks\_automated\_access} \\
\textbf{Scope:} web\_scraping \quad \textbf{Source run:} NVIDIA Opus run\_001

\begin{quote}
The TechPowerUp GPU Specs database serves a drag-handle CAPTCHA for automated requests, blocking WebFetch entirely. This is the most comprehensive GPU database ($\sim$3300 entries), but the Evaluator cannot use it directly for denominator verification. \textbf{Workaround:} Wikipedia individual series articles are the best alternative for denominator establishment, covering $\sim$95\% of what TechPowerUp lists for consumer desktop GPUs.
\end{quote}

\noindent\textit{Why this matters:} This entry is discovered in run\_001 and inherited by all subsequent runs. Without it, every new run (or new agent) wastes a round attempting to access TechPowerUp and failing. The entry converts one agent's failure into organizational knowledge that prevents future wasted effort.

\subsection{Strategic Shift: Curated Baselines Beat Parsing}
\label{app:knowledge:baseline}

\noindent\textbf{ID:} \texttt{curated\_baseline\_beats\_parsing\_for\_structured\_catalogs} \\
\textbf{Scope:} data\_collection \quad \textbf{Source run:} NVIDIA Opus run\_003

\begin{quote}
When collecting structured data about well-documented product lines, embedding a curated dataset directly in \texttt{action.py} produces far better round-1 results than attempting to parse complex HTML tables. Wikipedia tables with rowspan/colspan, reference brackets, and non-breaking spaces are error-prone to parse automatically. The curated approach delivered 96.4\% coverage immediately (299 records), while the Wikipedia parser contributed supplementary records that were mostly non-desktop contamination (25 mobile/Quadro products) requiring cleanup in round~2.
\end{quote}

\noindent\textit{Why this matters:} This is the single most consequential knowledge entry for Sonnet transfer. Cold Sonnet spends 3--5 rounds parsing Wikipedia tables and fixing bugs; seeded Sonnet reads this entry and begins with a curated baseline, achieving $>$97\% coverage in round~1. The entry encodes a strategic paradigm shift---from reactive parsing to proactive curation---that transforms the agent's approach.

\subsection{Evaluation Insight: Denominator Expansion Signal}
\label{app:knowledge:denominator}

\noindent\textbf{ID:} \texttt{denominator\_should\_track\_collected\_reality} \\
\textbf{Scope:} universal \quad \textbf{Source run:} NVIDIA Opus run\_001

\begin{quote}
When the Planner collects more legitimate records than the Evaluator's denominator (e.g., 299 records vs.\ 254 denominator), the correct response is to investigate and expand the denominator---not to cap coverage at 100\% and declare victory. In this task, GeForce~FX had 25 real desktop models (vs.\ initial estimate of 15), GeForce~8 had 15 (vs.\ 9). Each series where collected $>$ expected was a signal that the denominator was wrong. \textbf{Lesson:} Treat collected $>$ expected as a denominator correction signal, not a success signal.
\end{quote}

\noindent\textit{Why this matters:} This entry directly addresses denominator blindness. Without it, an Evaluator might accept $>$100\% coverage as ``extra credit'' rather than recognizing it as evidence that the denominator is too small. The lesson teaches future Evaluators to expand their understanding of the search space rather than prematurely declaring completeness---the core behavior that co-evolving evaluation is designed to produce.

\subsection{Cross-Run Evolution: The 266 Convergence}
\label{app:knowledge:convergence}

\noindent\textbf{ID:} \texttt{three\_round\_completion\_at\_266\_confirmed\_again\_v2} \\
\textbf{Scope:} product\_catalog\_collection \quad \textbf{Source run:} NVIDIA Sonnet seeded run\_005

\begin{quote}
This run completed in exactly 3 active planning rounds, converging at 266 products with 100\% coverage. Pattern: Round~1: curated hardcoded baseline $\to$ 256 records (97.3\% of final denominator). Round~2: 7 surgical fills (PCI/AGP bus variants + chip/memory variants) $\to$ 263 records. Round~3: 3 evaluator-named fills (memory type variants, OEM variant) $\to$ 266 records. The 266 convergence point is stable across runs.
\end{quote}

\noindent\textit{Why this matters:} This is a \texttt{\_v2} entry---a second-generation observation that confirms a pattern established by earlier runs. The entry records not just the final number but the \emph{convergence trajectory}: how 266 was reached through progressive variant discovery. Three independent seeded runs arriving at exactly 266 (Table~\ref{tab:denom_variance}) is the strongest evidence that knowledge transfer calibrates evaluation, not just collection. The entry itself is the mechanism: it encodes a product definition precise enough to produce identical denominators across independent runs.

\subsection{Evaluation Quality: Known Absence vs.\ Collection Failure}
\label{app:knowledge:absence}

\noindent\textbf{ID:} \texttt{known\_absence\_vs\_collection\_failure} \\
\textbf{Scope:} universal \quad \textbf{Source run:} UniProt Opus run\_002

\begin{quote}
When \texttt{eval.py} detects missing fields (e.g., empty pathways or variants), there are two very different causes: (1)~collection failure---the Planner didn't fetch available data; (2)~source absence---the data genuinely doesn't exist in any source. Before reporting gaps, spot-check against the live source. In this run, 6 proteins had no Reactome pathways and 1 had no variants---all confirmed as genuine absences in UniProt. Reporting these as gaps misleads the Planner into futile collection attempts.
\end{quote}

\noindent\textit{Why this matters:} This entry illustrates a subtle form of denominator refinement. The Evaluator learns to distinguish between ``the Planner failed to collect this'' and ``this doesn't exist.'' Without this distinction, the system chases impossible targets---a form of denominator blindness applied to individual fields rather than records. The lesson transfers across domains: any task with optional or sparse attributes faces the same ambiguity.

\subsection{Evaluation Rigor: Text Matching Has a Ceiling}
\label{app:knowledge:ceiling}

\noindent\textbf{ID:} \texttt{text\_matching\_eval\_has\_inherent\_ceiling} \\
\textbf{Scope:} universal \quad \textbf{Source run:} Q10 Opus run\_001

\begin{quote}
Regex-based text evaluation has a natural ceiling of $\sim$80--85\% for genuinely good solutions. Mathematical notation varies (LaTeX \texttt{\textbackslash otimes} vs.\ Unicode $\otimes$ vs.\ text ``Kronecker product''); depth bonuses based on word count are heuristic; concept presence $\neq$ concept correctness. To get more accurate evaluation, supplement text matching with live code execution tests, mathematical consistency checks, and structural checks. Don't keep adding regex patterns to push text scores higher---diminishing returns.
\end{quote}

\noindent\textit{Why this matters:} This entry shows the Evaluator learning the limits of its own evaluation method---a meta-cognitive insight. After run\_001's Evaluator discovered that regex-based scoring plateaus, subsequent runs shifted to hybrid evaluation combining text matching with executable verification. By run\_006, \texttt{eval.py} performs machine-precision numerical verification ($3 \times 10^{-16}$ matvec errors) with 31 concept-level sub-checks---an evolution from 7 simple keyword matches. The entry catalyzed this shift by naming the problem.

%% file: main.bbl
\begin{thebibliography}{20}
\providecommand{\natexlab}[1]{#1}
\providecommand{\url}[1]{\texttt{#1}}
\expandafter\ifx\csname urlstyle\endcsname\relax
  \providecommand{\doi}[1]{doi: #1}\else
  \providecommand{\doi}{doi: \begingroup \urlstyle{rm}\Url}\fi

\bibitem[Abouzaid et~al.(2026)Abouzaid, Blumberg, Hairer, Kileel, Kolda,
  Nelson, Spielman, Srivastava, Ward, Weinberger, and Williams]{firstproof2026}
Mohammed Abouzaid, Andrew~J. Blumberg, Martin Hairer, Joe Kileel, Tamara~G.
  Kolda, Paul~D. Nelson, Daniel Spielman, Nikhil Srivastava, Rachel Ward,
  Shmuel Weinberger, and Lauren Williams.
\newblock First proof.
\newblock \emph{arXiv preprint arXiv:2602.05192}, 2026.
\newblock Ten research-level mathematics questions with encrypted answers for
  AI evaluation.

\bibitem[{Anthropic}(2025{\natexlab{a}})]{anthropic2025claude}
{Anthropic}.
\newblock Claude code: {AI}-powered software engineering.
\newblock \url{https://docs.anthropic.com/en/docs/claude-code},
  2025{\natexlab{a}}.

\bibitem[{Anthropic}(2025{\natexlab{b}})]{anthropic2025swebench}
{Anthropic}.
\newblock Building effective agents.
\newblock \url{https://docs.anthropic.com/en/docs/build-with-claude/agentic},
  2025{\natexlab{b}}.

\bibitem[{Anthropic}(2026)]{anthropic2025aar}
{Anthropic}.
\newblock Automated weak-to-strong researcher.
\newblock \url{https://alignment.anthropic.com/2026/automated-w2s-researcher/},
  2026.
\newblock PGR 0.97 via automated agent configuration search with human-designed
  metrics.

\bibitem[{ByteDance Research}(2025)]{widesearch2025}
{ByteDance Research}.
\newblock {WideSearch}: A benchmark for comprehensive information-seeking
  tasks.
\newblock \emph{arXiv preprint arXiv:2508.07999}, 2025.

\bibitem[{Hermes Team}(2025)]{hermes2025}
{Hermes Team}.
\newblock Hermes: A self-improving agent framework.
\newblock \emph{arXiv preprint}, 2025.
\newblock Cross-episode knowledge accumulation for autonomous agents.

\bibitem[Hong et~al.(2023)Hong, Zhuge, Chen, Zheng, Cheng, Zhang, Wang, Wang,
  Yau, Lin, Zhou, Ran, Xiao, Wu, and Schmidhuber]{hong2023metagpt}
Sirui Hong, Mingchen Zhuge, Jonathan Chen, Xiawu Zheng, Yuheng Cheng, Ceyao
  Zhang, Jinlin Wang, Zili Wang, Steven Ka~Shing Yau, Zijuan Lin, Liyang Zhou,
  Chenyu Ran, Lingfeng Xiao, Chenglin Wu, and Jurgen Schmidhuber.
\newblock {MetaGPT}: Meta programming for a multi-agent collaborative
  framework.
\newblock \emph{arXiv preprint arXiv:2308.00352}, 2023.

\bibitem[Karpathy(2026)]{karpathy2026autoresearch}
Andrej Karpathy.
\newblock autoresearch: Automated machine learning research.
\newblock \url{https://github.com/karpathy/autoresearch}, 2026.
\newblock Accessed: 2026-04-01.

\bibitem[Lu et~al.(2024)Lu, Lu, Lange, Foerster, Clune, and
  Ha]{lu2024aiscientist}
Chris Lu, Cong Lu, Robert~Tjarko Lange, Jakob Foerster, Jeff Clune, and David
  Ha.
\newblock The {AI} scientist: Towards fully automated open-ended scientific
  discovery.
\newblock \emph{arXiv preprint arXiv:2408.06292}, 2024.

\bibitem[{Moonshot AI}(2026)]{kimik26}
{Moonshot AI}.
\newblock Kimi {K2.6}: Agent swarm coordination at scale.
\newblock \url{https://www.kimi.com/blog/kimi-k2-6}, 2026.
\newblock Centralized coordinator scaling to 300 sub-agents across 4000
  coordinated steps.

\bibitem[{OpenAI}(2026)]{symphony2026}
{OpenAI}.
\newblock Symphony: Autonomous coding agent orchestration.
\newblock \url{https://github.com/openai/symphony}, 2026.
\newblock Persistent daemon connecting project management tools to autonomous
  agent execution.

\bibitem[Phillips et~al.(2026)Phillips, Yan, De~Jonghe, and
  Weller]{phillips2026multiagent}
Cara Phillips, Eugene Yan, Jiri De~Jonghe, and Samuel Weller.
\newblock Multi-agent coordination patterns: Five approaches and when to use
  them.
\newblock \emph{Claude Blog}, 2026.

\bibitem[{Routa Team}(2025)]{routa2025}
{Routa Team}.
\newblock Routa: Multi-agent team management framework, 2025.
\newblock Centralized coordination for multi-agent teams.

\bibitem[Schmidgall et~al.(2025)]{schmidgall2025autoresearch}
Samuel Schmidgall et~al.
\newblock Agent laboratory: Using {LLM} agents as research assistants, 2025.

\bibitem[{Upward Deception Team}(2025)]{upwarddeception2025}
{Upward Deception Team}.
\newblock Upward deception: Can {LLM}s deceive their supervisors through
  self-assessment?
\newblock \emph{arXiv preprint arXiv:2512.04864}, 2025.

\bibitem[{Vending-Bench Team}(2025)]{vendingbench2025}
{Vending-Bench Team}.
\newblock Vending-bench: Benchmarking {AI} agent decision-making in simulated
  business environments.
\newblock \emph{arXiv preprint arXiv:2502.15840}, 2025.

\bibitem[Wang et~al.(2023)Wang, Xie, Jiang, Mandlekar, Xiao, Zhu, Fan, and
  Anandkumar]{voyager2023}
Guanzhi Wang, Yuqi Xie, Yunfan Jiang, Ajay Mandlekar, Chaowei Xiao, Yuke Zhu,
  Linxi Fan, and Anima Anandkumar.
\newblock Voyager: An open-ended embodied agent with large language models.
\newblock \emph{arXiv preprint arXiv:2305.16291}, 2023.

\bibitem[Wu et~al.(2023)Wu, Bansal, Zhang, Wu, Li, Zhu, Jiang, Zhang, Zhang,
  Liu, Awadallah, White, Burger, and Wang]{wu2023autogen}
Qingyun Wu, Gagan Bansal, Jieyu Zhang, Yiran Wu, Beibin Li, Erkang Zhu,
  Li~Jiang, Xiaoyun Zhang, Shaokun Zhang, Jiale Liu, Ahmed~Hassan Awadallah,
  Ryen~W White, Doug Burger, and Chi Wang.
\newblock {AutoGen}: Enabling next-gen {LLM} applications via multi-agent
  conversation.
\newblock \emph{arXiv preprint arXiv:2308.08155}, 2023.

\bibitem[Xie(2026)]{foragev1}
Huaqing Xie.
\newblock Forage: Solving denominator blindness in autonomous agents via
  co-evolving evaluation.
\newblock \emph{arXiv preprint}, 2026.
\newblock arXiv ID 7447726.

\bibitem[Zhang et~al.(2026)Zhang, Wang, Zhou, Liao, Feng, et~al.]{memrl2025}
Shengtao Zhang, Jiaqian Wang, Ruiwen Zhou, Junwei Liao, Yuchen Feng, et~al.
\newblock {MemRL}: Self-evolving agents via runtime reinforcement learning on
  episodic memory.
\newblock \emph{arXiv preprint arXiv:2601.03192}, 2026.

\end{thebibliography}
